\documentclass[fleqn,11pt]{article} 
\usepackage{hyperref}
\usepackage{url}
\usepackage{amsmath,amsthm,amssymb}
\usepackage{graphicx}
\usepackage{algorithm}
\usepackage{algorithmic}

\usepackage{caption}
\usepackage{subcaption}

\usepackage{wrapfig}
\usepackage{natbib}
\usepackage{fancyhdr}

\usepackage{tikz}
\usetikzlibrary{arrows}
\usetikzlibrary{positioning}
\definecolor {processblue}{cmyk}{0.96,0,0,0}

\DeclareMathOperator{\MMD}{MMD}

\begin{document}

\title{Removal of Batch Effects using Distribution-Matching Residual Networks}

\author{ 
Uri Shaham\,$^{1}$\thanks{The first two authors contributed equally to this work.} , 
Kelly P. Stanton\,$^{2}$\footnotemark[1] , 
Jun Zhao\,$^{3}$, 
Huamin Li\,$^{4}$, 
Khadir Raddassi$^{5}$,\\
Ruth Montgomery\,$^{6}$, 
and Yuval Kluger\,$^{2,3,4}$\thanks{To whom correspondence should be addressed, yuval.kluger@yale.edu}
}

\maketitle
\vspace{-3ex}
\begin{flushleft}
\small{
\hspace*{2ex}$^{1}$Department of Statistics, Yale University, New Haven, CT, USA\\
\hspace*{2ex}$^{2}$Department of Pathology, Yale School of Medicine, New Haven, CT, USA\\
\hspace{2ex}$^{3}$Program of Computational Biology and Bioinformatics, Yale University, New Haven, CT, USA\\
\hspace{2ex}$^{4}$Applied Mathematics Program, Yale University, New Haven, CT, USA\\
\hspace{2ex}$^{5}$Departments of Neurology and Immunobiology, Yale School of Medicine, New Haven, CT, USA\\
\hspace{2ex}$^{6}$Department of Internal Medicine, Yale School of Medicine, New Haven, CT, USA\\
}
\end{flushleft}

\begin{abstract}
Sources of variability in experimentally derived data include measurement error in addition to the physical phenomena of interest.  This measurement error is a combination of systematic components, originating from the measuring instrument, and random measurement errors.  Several novel biological technologies, such as mass cytometry and single-cell RNA-seq, are plagued with systematic errors that may severely affect statistical analysis if the data is not properly calibrated.
We propose a novel deep learning approach for removing systematic batch effects.  
Our method is based on a residual network, trained to minimize the Maximum Mean Discrepancy (MMD) between the multivariate distributions of two replicates, measured in different batches.
We apply our method to  mass cytometry and single-cell RNA-seq datasets, and demonstrate that it effectively attenuates batch effects.
\end{abstract}
%\begin{keywords}
%  Batch Effects, Calibration, Residual Networks, Maximum Mean Discrepancy, Deep Learning
%\end{keywords}

%%%%%%%%%%%%%%%%%%%%%%%%%%%%%%%%%%%%%%%%%%%%%%%%%%%%%%%%%%%%%%%%%%%%%%%%%%%%%%%%%%%%%%%%%%%%%%%%%%%%%%%
%%%%%%%%%%%%%%%%%%%%%%%%%%%%%%%%%%%%%%%%%%%%%%%%%%%%%%%%%%%%%%%%%%%%%%%%%%%%%%%%%%%%%%%%%%%%%%%%%%%%%%%

\section{Introduction}
Biological data are affected by the conditions of the measuring instruments. 
For example, biomedical data from replicated\footnote{We use the term \textit{replicates} to refer to technical replicates, i.e., multiple measurements of the same specimen, for example, two blood drops of the same person.} measurements, measured in different batches, may be distributed differently due to variation in these conditions between batches.
The term \textit{batch effects}, often used in the biological community, describes a situation where subsets (batches) of the measurements significantly differ in distribution, due to irrelevant instrument-related factors~\citep{leek2010tackling}. 
Batch effects introduce systematic error, which may cause statistical analysis to produce spurious results and/or obfuscate the signal of interest.

For example, CyTOF, a mass cytometry technique for measuring multiple protein levels in many cells of a biological specimen, is known to incur batch effects.  
When replicate blood specimens from the same patient are measured on a CyTOF machine in different batches (e.g. different days),  they might differ noticeably in the distribution of cells in the multivariate protein space.
In order to run a valid and effective statistical analysis on the data, a calibration process has to be carried out, to account for the effect of the difference in instrument conditions on the measurements. 

\begin{figure}
  \centering
  \includegraphics[width=3.0in]{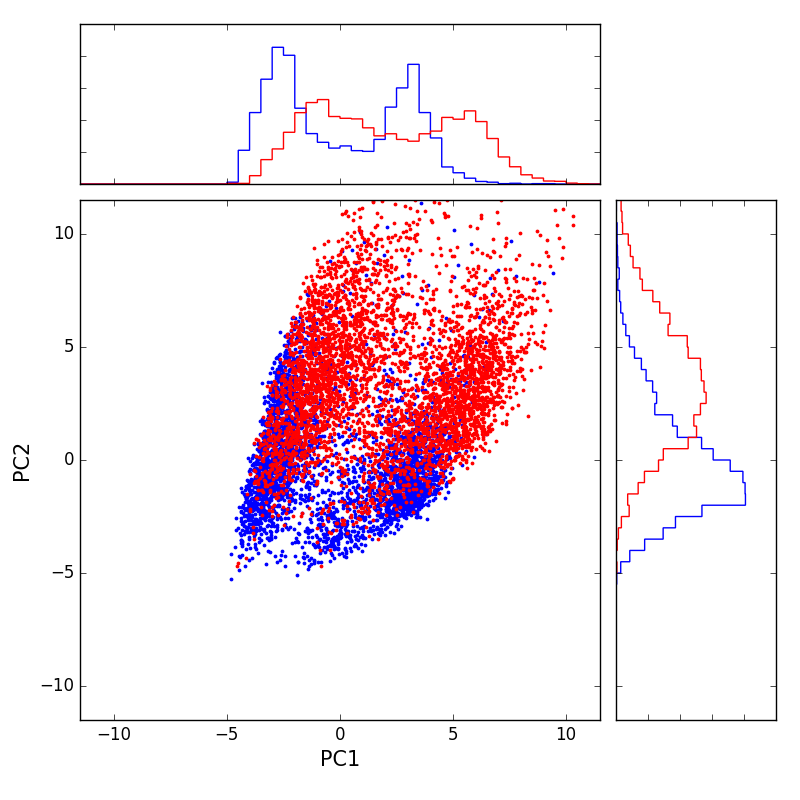}
  \includegraphics[width=3.0in]{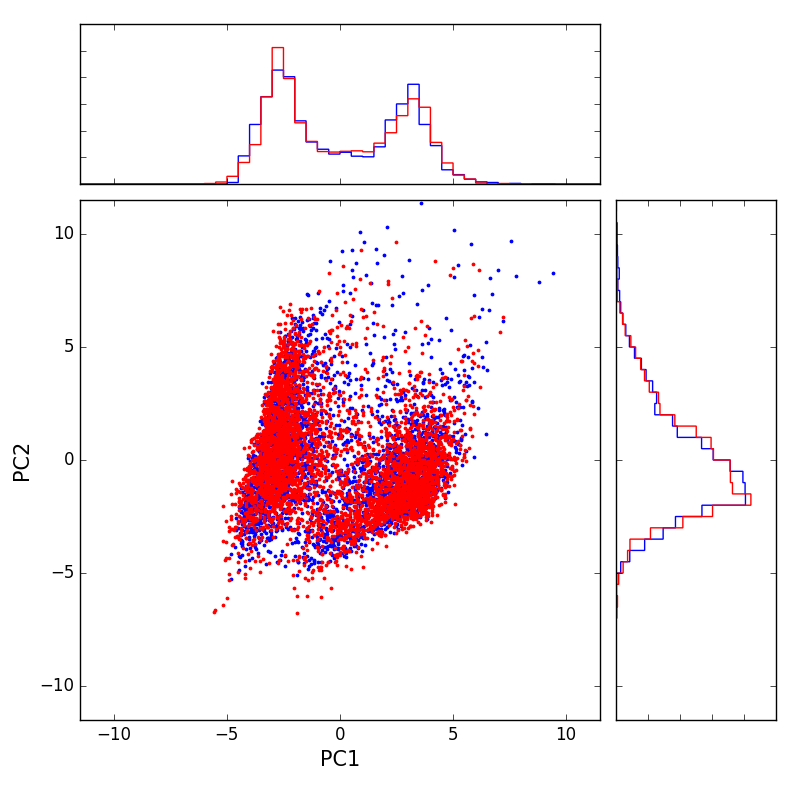}
  \caption{Calibration of CyTOF data. Projection of the source (red) and target (blue) samples on the first two principal components of the target data. Left: before calibration. Right: after calibration.}
 \label{fig:example}
 \end{figure}

Typically, the systematic effect of varying instrument conditions on the measurements depends on many unknown factors, whose impact on the difference between the observed and underlying true signal cannot be modeled.
In this manuscript, we consider cases where replicates differ in distribution, due to batch effects.
By designating one replicate to be the source sample\footnote{The term \textit{sample} is used with different meanings in the biological and statistical communities. Both meanings are used in this manuscript, however, usage should be clear from context. .} and the other to be the target sample, we propose a deep learning approach to learn a map that calibrates the distribution of the source sample to match that of the target. 
Our proposed approach is designed for data where the difference between these source and target distributions is moderate, so that the map that calibrates them is close to the identity map; such an assumption is fairly realistic in many situations.
An example of the problem and the output of our proposed method is depicted in Figure~\ref{fig:example}. 
A short demo movie is available at~\url{https://www.youtube.com/watch?v=Lqya9WDkZ60}.

To evaluate the effectiveness of our proposed approach, we employ it to analyze CyTOF and single-cell RNA-seq, and demonstrate that it successfully removes significant portions of the batch effect. We also demonstrate that it outperforms other popular approaches for calibration. To the best of our knowledge, similar performance on CyTOF data has never been reported. 

We justify our approach by demonstrating that shortcut connections are crucial to the success of calibration, as nets without shortcut connections might distort the biological properties of the data.
Furthermore, we provide evidence that a map from a source replicate to a target replicate, both extracted from the same specimen, can also be learned in an indirect manner, through maps between replicates from other specimens, without losing much accuracy. 
Generalizing this approach allows one to calibrate multiple source batches to a single target batch, where replicates from a single reference specimen are measured in each of these batches.

 The remainder of this manuscript is organized as follows: in Section~\ref{sec:preliminaries} we give a brief review of Maximum Mean Discrepancy and Residual Nets, on which our approach is based. The calibration learning problem is defined in Section~\ref{sec:calibration}, where we also describe our proposed approach. Experimental results on CyTOF and single-cell RNA-seq measurements are reported in Section~\ref{sec:experiments}. In Section~\ref{sec:relatedWork} we review some related works. In Section~\ref{sec:discussion} we discuss some technical aspects of our approach. Section~\ref{sec:conclusion} concludes the manuscript. 
 
%%%%%%%%%%%%%%%%%%%%%%%%%%%%%%%%%%%%%%%%%%%%%%%%%%%%%%%%%%%%%%%%%%%%%%%%%%%%%%%%%%%%%%%%%%%%%%%%%%%%%%%
%%%%%%%%%%%%%%%%%%%%%%%%%%%%%%%%%%%%%%%%%%%%%%%%%%%%%%%%%%%%%%%%%%%%%%%%%%%%%%%%%%%%%%%%%%%%%%%%%%%%%%%

\section{Preliminaries} \label{sec:preliminaries}

\subsection {Maximum Mean Discrepancy}
Maximum Mean Discrepancy (MMD,~\citet{gretton2012kernel, gretton2006kernel}) is a measure for distance between two probability distributions $p,q$. It is defined with respect to a function class $\mathcal{F}$ by 
\begin{equation}
\MMD(\mathcal{F}, p,q) \equiv \sup_{f \in \mathcal{F}}(\mathbb{E}_{x \sim p}f(x) - \mathbb{E}_{x \sim q}f(x)).\label{eq:MMD}
\end{equation}
When $\mathcal{F}$ is a reproducing kernel Hilbert space with kernel $k$, the $\MMD$ can be written as the distance between the mean embeddings of $p$ and $q$
\begin{equation}
\MMD^2(\mathcal{F}, p,q) = \|\mu_p - \mu_q \|^2_\mathcal{F}, \label{eq:meanEmb}
\end{equation}
where $\mu_p(t) = \mathbb{E}_{x\sim p} k(x,t)$. 
%the function $f$ which attains the supremum is
%\begin{equation}
%f(y) = \mathbb{E}_{x' \sim p}k(x,y) - \mathbb{E}_{x \sim q}k(x,y),\notag
%\end{equation}
Equation~\eqref{eq:meanEmb} can be written as
\begin{align}
\MMD^2(\mathcal{F}, p,q) = \mathbb{E}_{x,x' \sim p} k(x,x')-2\mathbb{E}_{x \sim p,y~\sim q} k(x,y) + \mathbb{E}_{y,y' \sim q} k(y,y'),\label{eq:mmd_k}
\end{align}
where $x$ and $x'$ are independent, and so are $y$ and $y'$. 
Importantly, if $k$ is a universal kernel, then $\text{MMD}(\mathcal{F}, p,q)=0$ iff $p=q$.
In practice, the distributions $p,q$ are unknown, and instead we are given observations $X=\{x_1,\ldots x_{n}\}, Y=\{y_1,\ldots y_{m}\}$, so that the (biased) sample version of~\eqref{eq:mmd_k} becomes
\begin{align}
&\MMD^2(\mathcal{F},X,Y) =\notag\\
&\frac{1}{n^2}\sum_{x_i, x_j \in X}k(x_i,x_j) - \frac{2}{nm}\sum_{x_i\in X, y_j \in Y}k(x_i,y_j) +\frac{1}{m^2}\sum_{y_i, y_j \in Y}k(y_i,y_j). \notag
\end{align}
MMD was originally proposed as a non-parametric two sample test, and has since been widely used in various applications. 
~\citet{li2015generative, dziugaite2015training}, use it as a loss function for neural net; here we adopt this direction to tackle the calibration problem, as discussed in Section~\ref{sec:calibration}.
%%%%%%%%%%%%%%%%%%%%%%%%%%%%%%%%%%%%%%%%%%%%%%%%%%%%%%%%%%%%%%%%%%%%%%%%%%%%%%%%%%%%%%%%%%%%%%%%%%%%%%%

\subsection {Residual Nets}
\begin{figure}[t]
\begin{center}
\begin{tikzpicture}[->,>=stealth',
  shorten >=1pt,
  node distance=1.05 cm and 2cm
  auto,
  main node/.style={rectangle,rounded corners,draw,align=center, text=blue, top color =white , bottom color = processblue!20}]
\node[main node] (1) {block input $x$};
\node[main node] (2) [above  of=1] {batch normalization};
\node[main node] (3) [above  of=2] {ReLU non-linearity};
\node[main node] (4) [above  of=3] {weight layer};
\node[main node] (5) [above  of=4] {batch normalization};
\node[main node] (6) [above  of=5] {ReLU non-linearity};
\node[main node] (7) [above  of=6] {weight layer};
\node[main node] (8) [above  of=7] {block output $x + \delta(x)$};

\path
(1) edge node [swap] {} (2)
(2) edge node [swap] {} (3)
(3) edge node [swap] {} (4)
(4) edge node [swap] {} (5)
(5) edge node [swap] {} (6)
(6) edge node [swap] {} (7)
(7) edge node [right=.1cm] { $\delta(x)$} (8)
(1) edge [bend right=90] node[right, midway]  {$x$} (8);
\end{tikzpicture}
\caption{A typical ResNet block.}
\label{fig:block}
\end{center}
\end{figure}
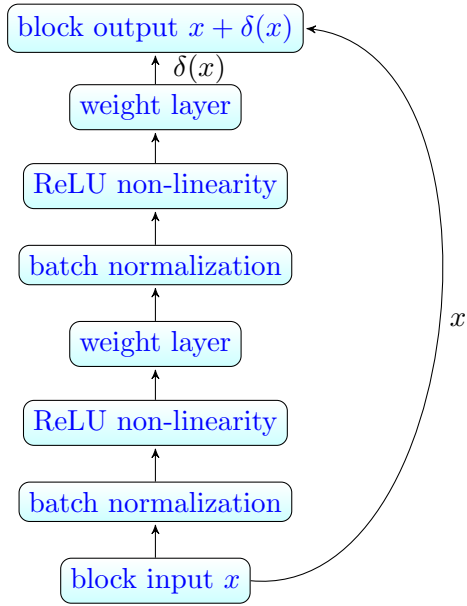

Residual neural networks (ResNets), proposed by~\citet{he2015deep} and improved in~\citep{he2016identity}, is a recently introduced class of very deep neural nets, mostly used for image recognition tasks. 
ResNets are typically formed by concatenation of many blocks, where each block receives an input $x$ (the output of the previous block) and computes output
$y = x + \delta(x)$,
where $\delta(x)$ is the output of a small neural net, which usually consists of two sequences of batch normalization~\citep{ioffe2015batch}, weight layers and non-linearity activations, as depicted in Figure~\ref{fig:block}.

It was empirically shown by~\citet{he2015deep} that the performance of very deep convolutional nets without shortcut connections deteriorates beyond some depth, while ResNets can grow very deep with increasing performance.
In a subsequent work,~\citet{he2016identity} showed that the gradient backpropagation in ResNets is improved, by avoiding exploding or vanishing gradients, comparing to networks without shortcut connections; this allows for more successful optimization, regardless of the depth.
~\citet{li2016demystifying} showed that ResNets with shortcut connections of depth 2 are easy to train, while deeper shortcut connections make the loss surface more flat. In addition, they argue that initializing ResNets with weights close to zero performs better than other standard initialization techniques.

Since a ResNet block consists of a residual term and an identity term, it can easily learn functions close to the identity function, when the weights are initialized close to zero, which is shown to be a valuable property for deep neural nets~\citep{hardt2016identity}.
In our case, the ability to efficiently learn functions which are close to the identity is appealing from an additional reason: 
we are interested in performing calibration between replicate samples whose multivariate distributions are close to each other;
to calibrate the samples, we are therefore interested in learning a map which is close to the identity map. 
A ResNet structure is hence a convenient tool to learn such a map. 

%%%%%%%%%%%%%%%%%%%%%%%%%%%%%%%%%%%%%%%%%%%%%%%%%%%%%%%%%%%%%%%%%%%%%%%%%%%%%%%%%%%%%%%%%%%%%%%%%%%%%%%
%%%%%%%%%%%%%%%%%%%%%%%%%%%%%%%%%%%%%%%%%%%%%%%%%%%%%%%%%%%%%%%%%%%%%%%%%%%%%%%%%%%%%%%%%%%%%%%%%%%%%%%

%%%%%%%%%%%%%%%%%%%%%%%%%%%%%%%%%%%%%%%%%%%%%%%%%%%%%%%%%%%%%%%%%%%%%%%%%%%%%%%%%%%%%%%%%%%%%%%%%%%%%%%
%%%%%%%%%%%%%%%%%%%%%%%%%%%%%%%%%%%%%%%%%%%%%%%%%%%%%%%%%%%%%%%%%%%%%%%%%%%%%%%%%%%%%%%%%%%%%%%%%%%%%%%

\section{Tackling the Calibration Problem} \label{sec:calibration}
Formally, we consider the following learning problem: let  $\mathcal{D}_1,\mathcal{D}_2$ be two distributions on $\mathbb{R}^d$, such that there exists a continuous map $\psi:\mathbb{R}^d \rightarrow \mathbb{R}^d$ so that if $X \sim \mathcal{D}_1$  then $\psi(X) \sim \mathcal{D}_2$. We also assume that $\psi$ is a small perturbation of the identity map. 

We are given two finite samples $\{x_1,\ldots, x_n\}, \{y_1,\ldots, y_m \}$ from $\mathcal{D}_1,\mathcal{D}_2$, respectively. 
The goal is to learn a map $\hat{\psi}:\mathbb{R}^d \rightarrow \mathbb{R}^d$ so that $\{\hat{\psi}(x_1),\ldots, \hat{\psi}(x_n)\}$ is likely to be a sample from $\mathcal{D}_2$.

Since we assume that $\psi$ is close to the identity, it is convenient to express it as $\psi(x) = x +\delta(x)$, where $\delta(x)$ is small, so that the connection to ResNets blocks becomes apparent. 

Our proposed solution, which we term MMD-ResNet is therefore a ResNet; the network gets two samples $\{x_1,\ldots x_n\}, \{y_1,\ldots, y_m \}$ of points in $\mathbb{R}^d$. We refer to $\{x_1,\ldots, x_n\}$ as the source sample and to $\{y_1,\ldots, y_m \}$ as the target sample. 
The net receives $\{x_1,\ldots x_n\}$ as input and is trained to learn a map of the source sample, to make it similar in distribution to the target sample.
Specifically, we train the net with the following loss function
\begin{equation}
L(w) = \sqrt{\MMD^2(\{\hat{\psi}(x_1),\ldots \hat{\psi}(x_n)\}, \{y_1,\ldots y_m \})}, \notag
\end{equation}
where $\hat{\psi}$ is the map computed by the network, and depends on the network parameters $w$. We train the net in a stochastic mode, so that in fact the $\MMD$ is computed only on mini-batches from both samples, and not on the entire samples.

%%%%%%%%%%%%%%%%%%%%%%%%%%%%%%%%%%%%%%%%%%%%%%%%%%%%%%%%%%%%%%%%%%%%%%%%%%%%%%%%%%%%%%%%%%%%%%%%%%%%%%%
%%%%%%%%%%%%%%%%%%%%%%%%%%%%%%%%%%%%%%%%%%%%%%%%%%%%%%%%%%%%%%%%%%%%%%%%%%%%%%%%%%%%%%%%%%%%%%%%%%%%%%%

\section{Experimental Results} \label{sec:experiments}
In this section we report experimental results on biological data obtained using two types of high-throughput technologies: CyTOF and single-cell RNA-seq (scRNA-seq). 
CyTOF is a mass cytometry technology that allows simultaneous measurements of multiple protein markers in each cell of a specimen (e.g., a blood sample), consisting of $10^4-10^6$ cells~\citep{spitzer2016mass}. 
scRNA-seq is a sequencing technology that allows to simultaneously measure mRNA expression levels of all genes in thousands of single cells.
%%%%%%%%%%%%%%%%%%%%%%%%%%%%%%%%%%%%%%%%%%%%%%%%%%%%%%%%%%%%%%%%%%%%%%%%%%%%%%%%%%%%%%%%%%%%%%%%%%%%%%%
\subsection{Technical Details}
All MMD-ResNets were trained using RMSprop~\citep{tieleman2012lecture}, using the Keras default hyper-parameter setting; a penalty of 0.01 on the $l_2$ norm of the network weights was added to the loss for regularization. We used mini-batches of size 1000 from both the source and target samples. 
A subset 10\% of the training data was held out for validation, to determine when to stop the training.

The kernel we used is a sum of three Gaussian kernels
\begin{equation}
k(x,y) = \sum_i \exp\left(-\frac{\|x-y\|^2}{\sigma_i^2}\right). \notag
\end{equation}
We chose the $\sigma_i$s to be $\frac{m}{2}, m, 2m$, where $m$ is the median of the average distance between a point in the target sample to its nearest 25 neighbors.

We implemented our net in Keras; our codes and data are publicly available at~\url{https://github.com/ushaham/BatchEffectRemoval.git}. 
%%%%%%%%%%%%%%%%%%%%%%%%%%%%%%%%%%%%%%%%%%%%%%%%%%%%%%%%%%%%%%%%%%%%%%%%%%%%%%%%%%%%%%%%%%%%%%%%%%%%%%%

\subsection{Calibration of CyTOF Data}\label{sec:cytof}
Mass cytometry uses a set of antibodies, each of which is conjugated to a unique heavy ion and binds to a different cellular protein.  Cells are then individually nebulized and subjected to mass spectrometry.  
Protein abundance is indirectly observed from the signal intensity at each protein's associated ions' mass to charge ratio. Multiple specimens can be run in the same batch by using barcoding with additional ions to record the origin of each specimen~\citep{spitzer2016mass}.
A CyTOF batch contains measurements of numerous cells from a few specimens, and each batch is affected by systematic errors~\citep{finck2013normalization}. 
%%%%%%%%%%%%%%%%%%%%%%%%%%%%%%%%%%%%%%%%%%%%%%%%%%%%%%%%%%%%%%%%%%%%%%%%%%%%%%%%%%%%%%%%%%%%%%%%%%%%%%%

\subsubsection{Data}\label{sec:cytofData}
Our calibration experiments were performed on data collected at Yale New Haven Hospital; Peripheral Blood Mononuclear Cells (PBMC) were collected from two MS patients at baseline and 90 days after Gilenya treatment and cryopreserved. At the end of the study PBMC were thawed in two batches (on two different days) and incubated with or without PMA+ionomycin (using a robotic platform). PMA/ionomycin stimulated and unstimulated samples were barcoded using Cell-ID (Fluidigm), then pooled and labeled for different markers with mass cytometry antibodies and analyzed on CyTOF III Helios.  
%the data was generated from two specimens of two patients that were measured on the same CyTOF instrument in two different days. Each patient had two specimens in two different biological conditions - baseline, and 3 months of treatment. 
Altogether we used a collection containing eight samples: 2 patients $\times$ 2 conditions $\times$ 2 days. From this collection, we assembled four source-target pairs, where for each patient and biological condition, the sample from day 1 was considered as source and the one from day 2 as target. 
All samples were of dimension $d=25$ \footnote{See a full specification of the markers in Appendix~\ref{app:markers}} and contained 1800-5000 cells. 
%%%%%%%%%%%%%%%%%%%%%%%%%%%%%%%%%%%%%%%%%%%%%%%%%%%%%%%%%%%%%%%%%%%%%%%%%%%%%%%%%%%%%%%%%%%%%%%%%%%%%%%

\subsubsection{Pre-processing}\label{sec:pp}
All samples were manually filtered by a human expert to remove debris and dead cells. Log transformation, a standard practice for CyTOF, was applied to the data.
In addition, a bead-normalization procedure was applied to the data; this is a current practice for normalizing CyTOF data~\citep{finck2013normalization}. Yet, our results demonstrate that the samples clearly differ in distribution, despite the fact that they were normalized. 

A typical CyTOF sample contains large proportions of zero values (up to 40\% sometimes) which occur due to instabilities of the CyTOF instrument and usually do not reflect biological phenomenon. 
As leaving the zero values in place might incur difficulties to calibrate the data, a cleaning procedure has to be carried out. 
In our experiments we collected the cells with non or very few zero values and used them to train a denoising autoencoder(DAE~\citet{vincent2008extracting}). 
Specifically, the DAE was trained to reconstruct clean cells $x$ from noisy inputs $\tilde{x}$, where the $\tilde{x}$ was obtained from $x$ by multiplying each entry of $x$ by a independent Bernoulli random variable with parameter $=0.8$. 
The DAE contained two hidden layers, each of 25 ReLU units; the output units were linear. As with the MMD-ResNets, the DAEs were also trained using RMSprop, and their loss contained $l_2$ penalization of the weights.
Once the DAE was trained, we passed the source and target samples through it, and used their reconstructions, which did not contain zeros, for the calibration. 
In all our CyTOF experiments, source and target refer to the denoised version of these samples.
Lastly, as a standard practice, in each of the experiment the input to the net (i.e., the source sample) was standardized to have zero mean and unit variance in each dimension. The parameters of the standardization were then also applied to the target sample.

%%%%%%%%%%%%%%%%%%%%%%%%%%%%%%%%%%%%%%%%%%%%%%%%%%%%%%%%%%%%%%%%%%%%%%%%%%%%%%%%%%%%%%%%%%%%%%%%%%%%%%%
\subsubsection{CyTOF calibration}\label{sec:cytofCali}
We trained a MMD-ResNet on each of the four source-target pairs. All nets were identical, and contained three blocks, where each block is as in Figure~\ref{fig:block}. Each of the weight matrices was of size $25 \times 25$. The net weights were initialized by sampling from a $\mathcal{N}(0,10^{-4})$ distribution.
The projection of the target and source data onto the first two PCs of the target sample in a representative source-target pair is shown in Figure~\ref{fig:example}. The plots of the remaining three pairs are presented in Appendix~\ref{app:other3}.
In the left plot, it is apparent that before calibration, the source sample (red) differs in distribution from the target sample (blue). After calibration (right plot), the gap between the source and the target distributions decreases significantly.
The MMD between the source and target before and after calibration in each of the four pairs is shown in Table~\ref{tab:cytof_mmd}, in addition to the MMD obtained using a multi-layer perceptron (MLP) MMD-net with a similar architecture to the ResNet, except without shortcut connections. The MLP was initialized in a standard fashion~\citep{glorot2010understanding}.
\begin{table}[t]
\centering
\caption {CyTOF calibration experiment: $\MMD$ values between random batches of size 1000 from the source and target samples, before and after calibration on each of the four source-target pairs (patient1-baseline, patient2-baseline, patient1-treatment, patient2-treatment). The $\MMD$ between two random batches of the target sample is provided as reference in the bottom row. The calibrated data is significantly closer in $\MMD$ to the target sample. The presented values are average$\pm$std, based on sampling of five random subsets of size 1000.}
\vskip 0.15in
\begin{tabular}{ l || c | c  | c | c}
  \hline                        
  MMD to target \textbackslash pair  & pa.1 base. & pa.2 base. & pa.1 treat. & pa.2 treat. \\ \hline\hline      
  no calibration  & 0.66$\pm$0.01 & 0.56$\pm$0.01 & 0.59$\pm$0.01 & 0.70$\pm$0.01 \\ \hline
  MLP calibration  & 0.55$\pm$0.01 & 0.18$\pm$0.01 & 0.26$\pm$0.01 & 0.21$\pm$0.01 \\ \hline
  ResNet calibration  & 0.27$\pm$0.01 & 0.17$\pm$0.01 & 0.24$\pm$0.01 & 0.17$\pm$0.01 \\ \hline\hline      
  MMD(target,target)  & 0.12$\pm$0.01 & 0.12$\pm$0.01 & 0.13$\pm$0.01 & 0.13$\pm$0.01 \\ \hline
  \end{tabular} 
  \label{tab:cytof_mmd}
\end{table}
As can be seen, the calibrated data is significantly closer to the target data than the original source data. 
The ResNet achieves similar performance to the MLP on two pairs and outperforms the MLP on the other two. In Section~\ref{sec:cytof_valid} we will show that ResNet architecture is in fact a crucial element in our approach, for a more important reason.

\begin{figure*}[t!]
  \centering
  \includegraphics[width=2.8in]{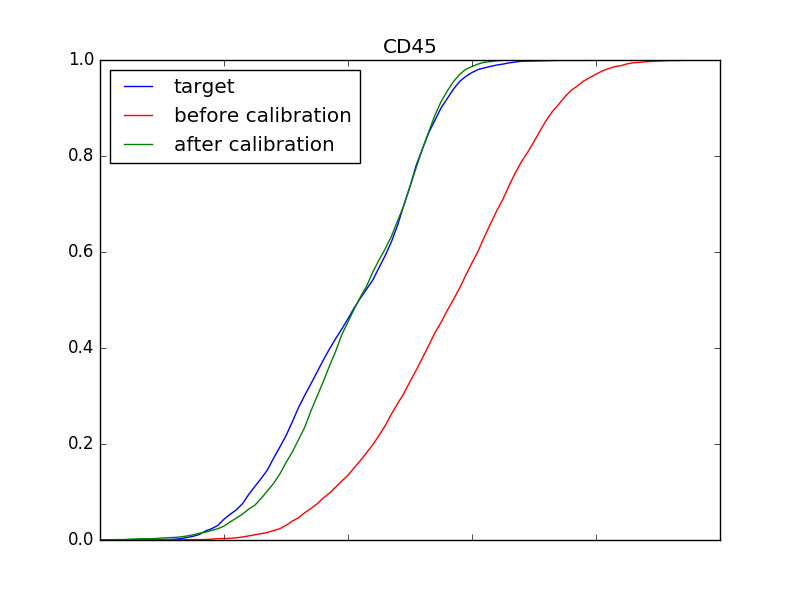}
  \includegraphics[width=2.8in]{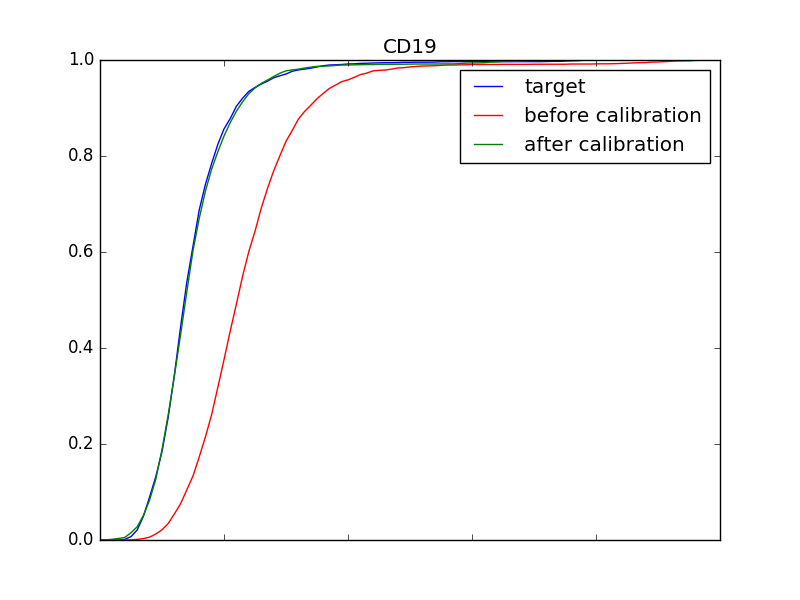}
  \includegraphics[width=2.8in]{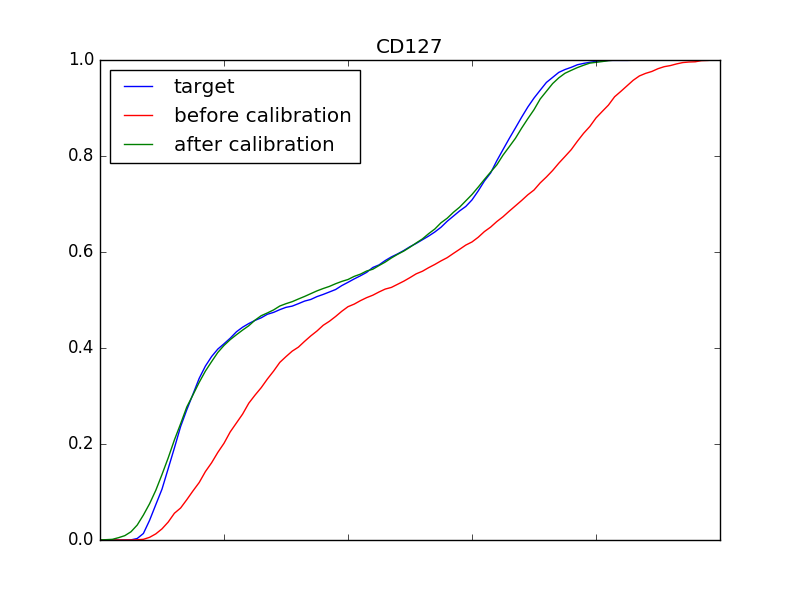}
  \includegraphics[width=2.8in]{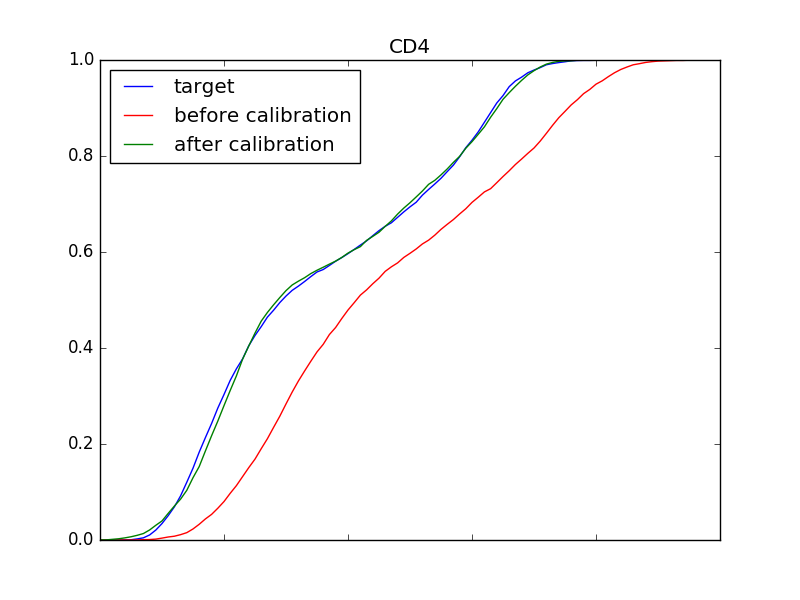}
  \includegraphics[width=2.8in]{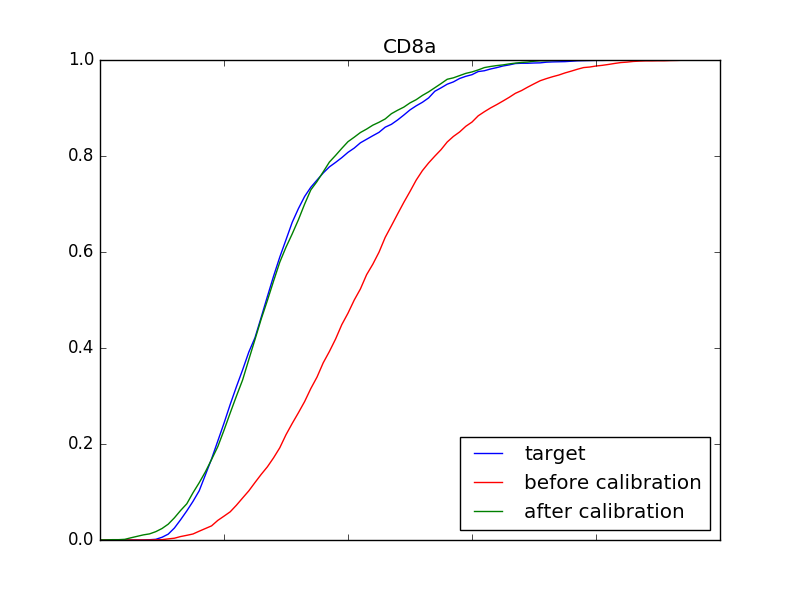}
  \includegraphics[width=2.8in]{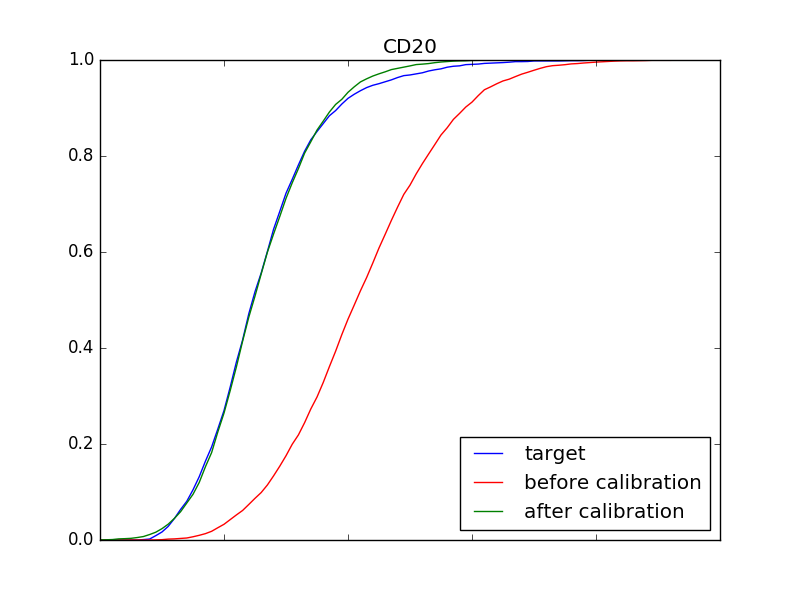}
  \caption{A marginal perspective on the quality of calibration. Empirical cumulative distribution functions of the first six markers in the CyTOF calibration experiment. In each plot the blue, red and green curves corresponds to the target, source and calibrated source samples, respectively. In each marker the blue and green curves are substantially closer than the blue and red curves.}
 \label{fig:markers}
 \end{figure*}

On a per-marker level, Figure~\ref{fig:markers} shows the empirical cumulative distribution functions of the first six markers in the source sample before and after the calibration, in comparison to the target sample. In all cases, as well as on the remaining markers that are not shown here, the calibrated source curves are substantially closer to the target than the  curves before calibration.

%%%%%%%%%%%%%%%%%%%%%%%%%%%%%%%%%%%%%%%%%%%%%%%%%%%%%%%%%%%%%%%%%%%%%%%%%%%%%%%%%%%%%%%%%%%%%%%%%%%%%%%

\subsubsection{Biological Validation and the Importance of Shortcut Connections}\label{sec:cytof_valid}
\begin{figure}
  \centering
  \includegraphics[width=2.0in]{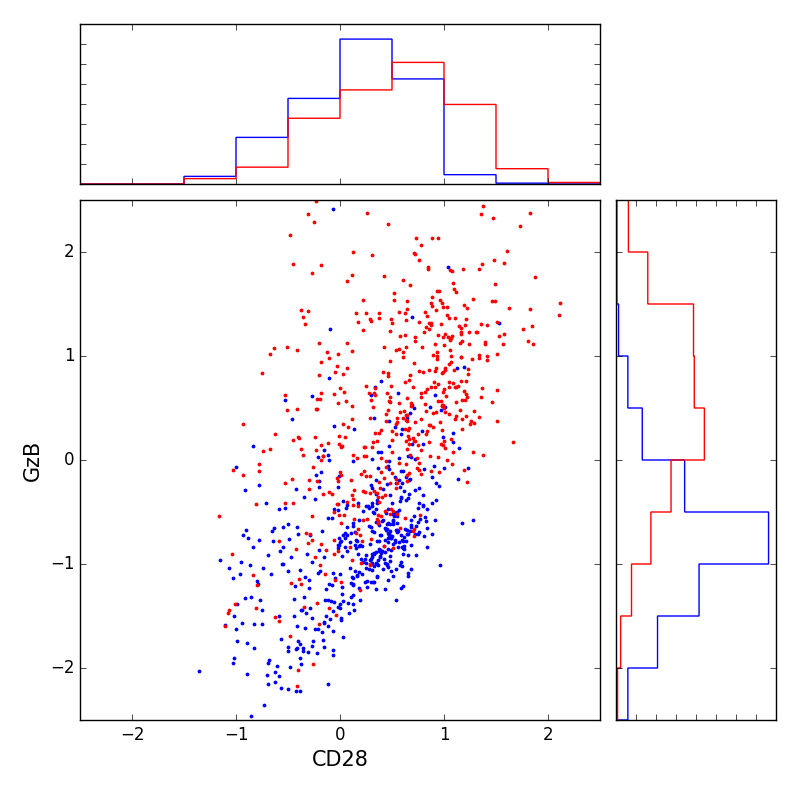}
  \includegraphics[width=2.0in]{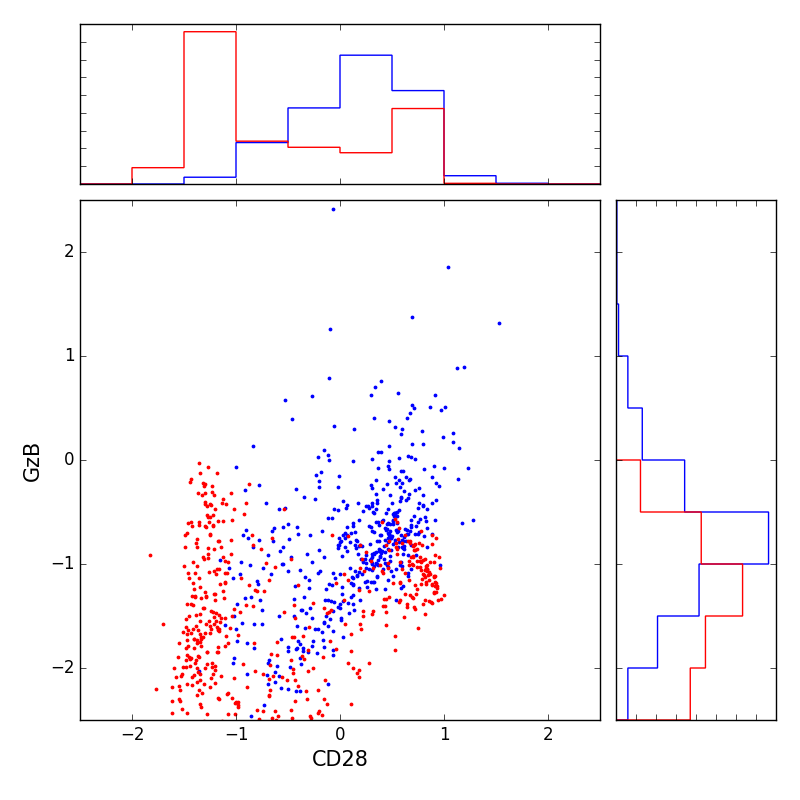}
  \includegraphics[width=2.0in]{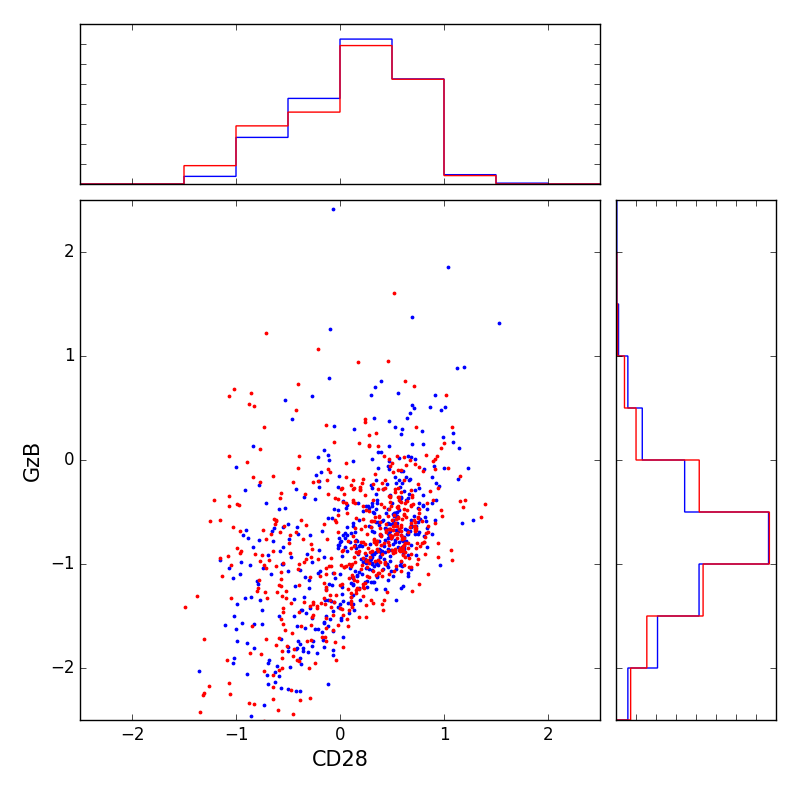}
  \caption{Calibration of CyTOF data: CD8+T-cells (red) and target (blue) samples in the (CD28, GzB) plane. Left: before calibration. Center: calibration using MLP. Right: calibration using ResNet.}
 \label{fig:cd8}
 \end{figure}

To biologically assess the quality of the calibration and further justify our proposed approach, we inspect the effect of calibration not only at a global level across all types of cell sub-populations, but also zoom in to a specific cell sub-population.
Specifically, we focus here on CD8+T-cells, also known as Killer T-cells, in the 2D space of the markers CD28 and GzB. 
In each sample, we identified the CD8+T-cells sub-population based on manual gating, performed by a human expert.
Figure~\ref{fig:cd8} shows the CD8+T-cells of the source and target samples from the baseline samples of patient 2 (patient2-baseline), before calibration, after calibration using ResNet and after calibration using similar net without shortcut connections (MLP). 
As can be seen, when the calibration is performed by a net without shortcut connections, the CD8+T-cells sub-population is not mapped to the same region as its target sample counterpart. However, with ResNet it is mapped appropriately.

The MMD score between the target sample and the ResNet-calibrated source sample was very similar to the MMD score between the target sample and the MLP-calibrated source sample. 
We therefore see that in order to achieve good calibration it does not suffice that the calibrated source sample will be close in MMD to the target sample. It is also crucial that the calibration map will be close to the identity. Nets without shortcut connections can clearly compute maps which are close to the identity. However, when trained to minimize MMD, the resulting map is not necessarily close to the identity, as there might be different maps that yield low MMD, despite being far from the identity, and are easier to reach from random initialization by optimization. Therefore, to obtain a map that is close to the identity, ResNet is a more appropriate tool, if not crucial, comparing to nets without shortcut connections.

The plots for the remaining three source-target pairs are shown in Appendix~\ref{app:other3}.
%%%%%%%%%%%%%%%%%%%%%%%%%%%%%%%%%%%%%%%%%%%%%%%%%%%%%%%%%%%%%%%%%%%%%%%%%%%%%%%%%%%%%%%%%%%%%%%%%%%%%%%

\subsubsection{Comparison to Linear Methods}\label{sec:linear}
In this section we compare the quality of calibration of our MMD-ResNet to two of the most popular techniques for removal of batch effects.
The simplest and most common~\citep{nygaard2016methods} adjustment is zero centering, i.e., substracting from any value the global mean of its batch; see, for example the \textit{batchadjust} command in the R package PAMR~\citep{hastie2015package}. 
The first linear method that we consider here is calibration by matching each marker's mean and variance in the source sample to the corresponding values in target sample.

The second common practice is to obtain the principal components of the data, and remove the components that are most correlated with the batch index~\citep{liu2016evaluation}. 
Table~\ref{tab:linear} compares the performances of our approach and the two approaches mentioned above in terms of MMD scores.
As can be seen, the calibration obtained from our MMD-ResNet outperforms the ones obtained by other two methods.
 \begin{table}[t]
\centering
\caption {CyTOF calibration: Comparison of calibration using (1) matching mean and variance of each marker, (2) PCA and (3) MMD-ResNet. The table entrees are average $\MMD$ between the target sample and the calibrated source sample, based on five random subsets of size 1000.}
\vskip 0.15in
\begin{tabular}{ l || c | c  | c | c}
  \hline                        
  MMD to target \textbackslash pair  & pa.1 base. & pa.2 base. & pa.1 treat. & pa.2 treat. \\ \hline\hline      
  mean, var. matching  & 0.26$\pm$0.02 & 0.25$\pm$0.01 & 0.30$\pm$0.01 & 0.30$\pm$0.02 \\ \hline
  PCA                  & 0.38$\pm$0.02 & 0.39$\pm$0.01 & 0.44$\pm$0.01 & 0.37$\pm$0.01 \\ \hline
  ResNet               & 0.27$\pm$0.01 & 0.18$\pm$0.01 & 0.24$\pm$0.01 & 0.17$\pm$0.01 \\ \hline

  \end{tabular} 
  \label{tab:linear}
\end{table}
%%%%%%%%%%%%%%%%%%%%%%%%%%%%%%%%%%%%%%%%%%%%%%%%%%%%%%%%%%%%%%%%%%%%%%%%%%%%%%%%%%%%%%%%%%%%%%%%%%%%%%%
%%%%%%%%%%%%%%%%%%%%%%%%%%%%%%%%%%%%%%%%%%%%%%%%%%%%%%%%%%%%%%%%%%%%%%%%%%%%%%%%%%%%%%%%%%%%%%%%%%%%%%%

\subsection{Calibration of Single-Cell RNA-seq Data}\label{sec:rna}

Drop-seq~\citep{macosko2015highly} is a novel technique for simultaneous measurement of single-cell mRNA expression levels of all genes of numerous individual cells. 
Unlike traditional single cell sequencing methods, which can only sequence up to hundreds or a few thousands of cells~\citep{picelli2013smart}, \citep{jaitin2014massively},
Drop-seq enables researchers to analyze many thousands of cells in parallel, thus offers a better understanding of complex cell types of cellular states.

However, even with several thousands of cells ($\sim$5000) in each run, only less than half of the cells typically contain enough transcribed genes, that can be used for statistical analysis. 
As the number of cells in a single run is not sufficient for studying very complicated tissues, one needs to perform multiple runs, in several batches, so that the cumulative number of cells is a good representation of the distribution of cell populations. 
This process may create batch effects, which need to be removed.

In~\citep{shekhar2016comprehensive}, seven replicates from two batches were sequenced to study bipolar cells of mouse retina.
Applying their approach to clean and filter the data, we obtained a dataset of 13,166 genes, each expressed in more than 30 cells and has a total transcript count of more than 60, and 27,499 cells, each of which has more than 500 expressed genes.
Data was then normalized such that counts in each cell sum to 10000, followed by a $\log$ transform of (count + 1).
~\citet{shekhar2016comprehensive} estimated that most of the signal is captured by the leading 37 principal components and used them for downstream analysis. We therefore projected the 13,166-dimensional data onto the subspace of the first 37 principal components and used this reduced data for our calibration experiment.

\begin{figure}
  \centering
  \includegraphics[width=3.0in]{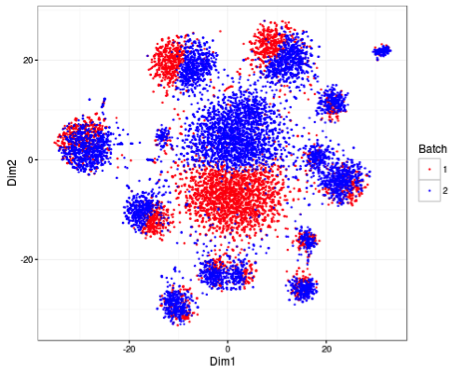}
  \includegraphics[width=3.0in]{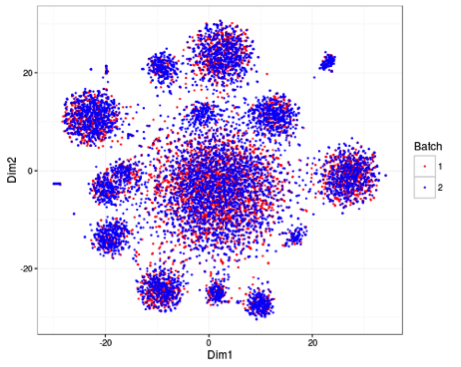}
  \caption{Calibration of scRNA-seq. $t$-SNE plots before (left) and after (right) calibration using MMD-ResNet.}
 \label{fig:RNA}
 \end{figure}

We arbitrarily chose batch 1 to be the target and the one from batch 2 to be the source, and used them to train a MMD-ResNet. The net had three blocks, where each block is as in Figure~\ref{fig:block}. In each block, the two weight matrices were of size 37 $\times$ 50 and 50 $\times$ 37. The net weights were initialized by sampling from a $\mathcal{N}(0,10^{-4})$ distribution.
$t$-SNE plots of the data before and after calibration are presented in Figure~\ref{fig:RNA}, which shows that after calibration, clusters from the source batches are mapped onto their target batch counterparts. 

Table~\ref{tab:rna} shows the MMD between the source and target batch before and after calibration, in comparison to the two linear calibration methods mentioned in Section~\ref{sec:linear}, as well as to Combat~\citep{johnson2007adjusting}, a standard technique for batch effect removal, which performs linear adjustments, where the corrections are based on Bayesian estimation. Combat and the mean-variance matching were applied on the full set of 13,166 genes, after normalization as in~\citep{shekhar2016comprehensive}, rather than on the projection of the data onto the leading 37 principal components, which was the input to the MMD-ResNet on this dataset.
 \begin{table}[t]
\centering
\caption {RNA calibration. Comparison of calibration using (1) matching mean and variance of each gene, (2) PCA, (3) Combat and (4) MMD-ResNet. The table entrees are average $\MMD$ between the target sample and the calibrated source sample, based of five random subsets of size 1000. The $\MMD$ between two random batches of the target sample is provided as reference in the rightmost column.}
\vskip 0.15in
\begin{tabular}{ c | c | c  | c | c || c}
  \hline                        
  before calib. & mean, var. matching & PCA & Combat & ResNet  & target-target\\ \hline\hline      
    0.43 $\pm$0.01 & 0.25$\pm$0.01 & 0.21$\pm$0.01 & 0.15$\pm$0.01 & 0.12$\pm$0.01 & 0.11
  \end{tabular} 
  \label{tab:rna}
\end{table}
As can be seen, MMD-ResNet outperforms all other methods in terms of MMD. 

\begin{figure}
  \centering
  \includegraphics[width=2.0in]{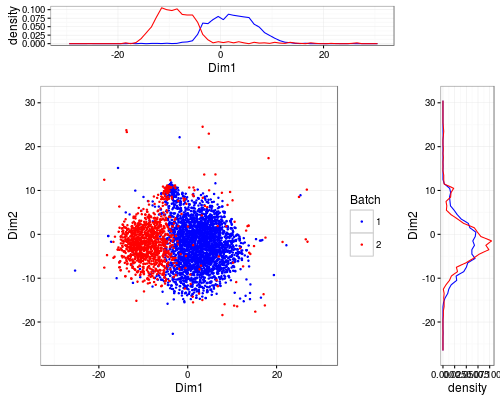}
  \includegraphics[width=2.0in]{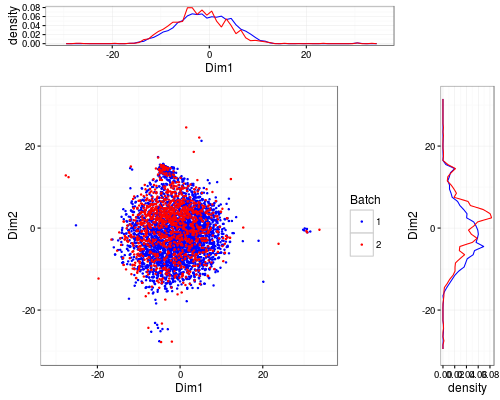}
  \includegraphics[width=2.0in]{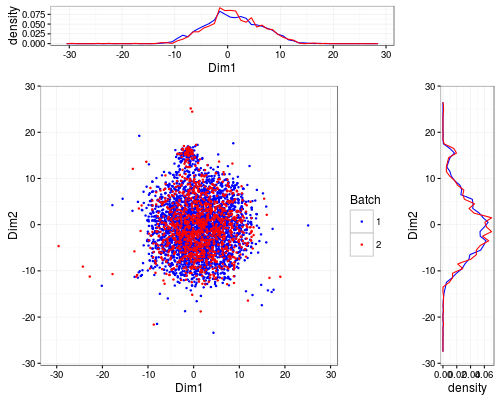}
  \caption{Calibration of cells with  high expression of Prkca. $t$-SNE plots before calibration (left), after calibration using Combat (middle) and MMD-ResNet (right).}
 \label{fig:Prkca}
 \end{figure}

To further assess the quality of calibration, and verify that our approach does not distort the underlying biological patterns in the data, we examine the sub-population of cells with high log-transformed expression values ($\ge 3$) of the \textit{Prkca} marker (which characterizes the cell sub-population of the large cluster in Figure~\ref{fig:RNA}). Figure~\ref{fig:Prkca} shows this sub-population before and after calibration, as well as after calibration using Combat. As can be seen, this sub-population is calibrated appropriately. Visually, in this analysis, MMD-ResNet achieves better calibration than Combat.
%%%%%%%%%%%%%%%%%%%%%%%%%%%%%%%%%%%%%%%%%%%%%%%%%%%%%%%%%%%%%%%%%%%%%%%%%%%%%%%%%%%%%%%%%%%%%%%%%%%%%%%

\subsection{Indirect Calibration}\label{sec:indirect}
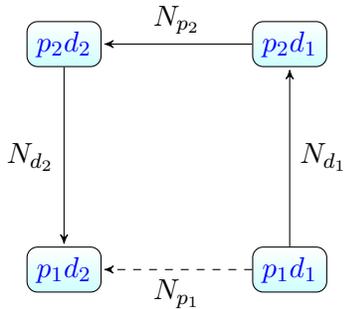
\begin{figure}[t]
\begin{center}
\begin{tikzpicture}[->,>=stealth',
  shorten >=1pt,
  node distance=3cm and 3cm
  auto,
  main node/.style={rectangle,rounded corners,draw,align=center, text=blue, top color =white , bottom color = processblue!20}]
\node[main node] (1) {$p_1d_1$};
\node[main node] (2) [above  of=1] {$p_2d_1$};
\node[main node] (3) [left  of=1] {$p_1d_2$};
\node[main node] (4) [left  of=2] {$p_2d_2$};

\path
(1) edge node [right, midway]  {$N_{d_1}$} (2)
(2) edge node [above, midway] {$N_{p_2}$} (4)
(4) edge node [left, midway] {$N_{d_2}$} (3)
(1) edge [dashed] node [below, midway] {$N_{p_1}$} (3);
\end{tikzpicture}
\caption{Indirect calibration experiment scheme.}
\label{fig:indirectScheme}
\end{center}
\end{figure}

In this section we demonstrate how MMD-ResNets can be used to calibrate a source distribution to a target distribution in an indirect manner, i.e., without training a net to learn this map directly, as in the previous experiments. 
For this experiment we use four of the CyTOF samples described in Section~\ref{sec:cytofData}. i.e., samples from patients 1 and 2 at baseline condition, each measured on the instrument in day 1 and day 2. 
We use the shorthand notation $p_1d_1$ to refer to the sample of patient 1 measured in day 1 and similarly $p_1d_2,p_2d_1,p_2d_2$ to the other samples.
In Section~\ref{sec:cytofCali} we trained a MMD-ResNet (which we now denote by $N_{p_1}$) that maps $p_1d_1$ to $p_1d_2$ and a ResNet $N_{p_2}$ which maps $p_2d_1$ to $p_2d_2$. In the following experiment we will map $p_1d_1$ to $p_1d_2$ indirectly. 
The setup is as follows:
In addition to the nets $N_{p_1}$,$N_{p_2}$ that were trained in Section~\ref{sec:cytofCali}, we train two additional MMD-ResNets, a ResNet $N_{d_1}$, mapping $p_1d_1$ to $p_2d_1$ and a ResNet $N_{d_2}$, mapping $p_2d_2$ to $p_1d_2$.
A scheme showing direct and indirect calibrations is shown in Figure~\ref{fig:indirectScheme}.

\begin{figure}
  \centering
  \includegraphics[width=2.0in]{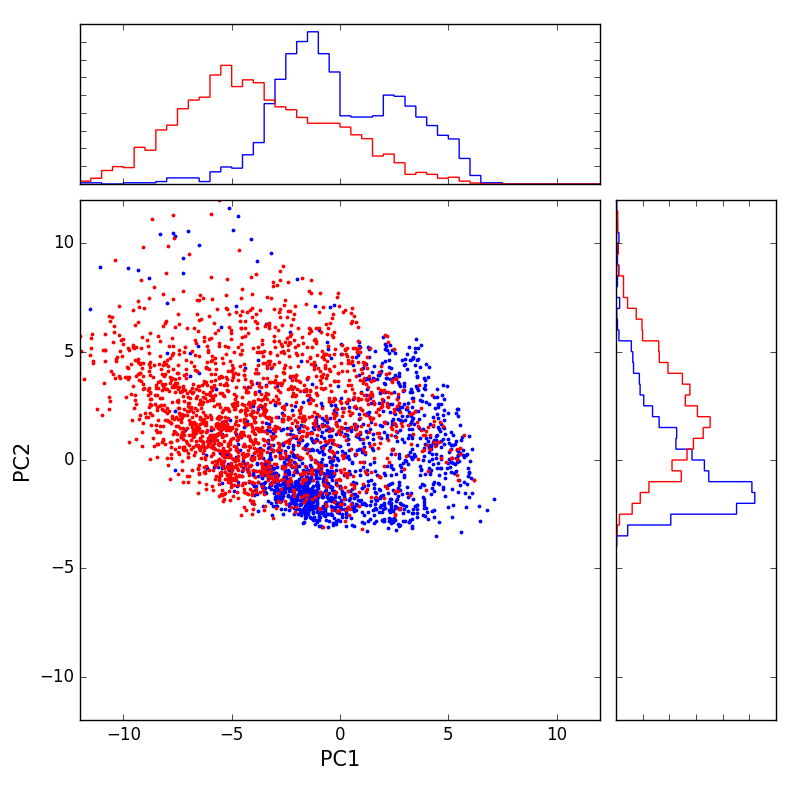}
  \includegraphics[width=2.0in]{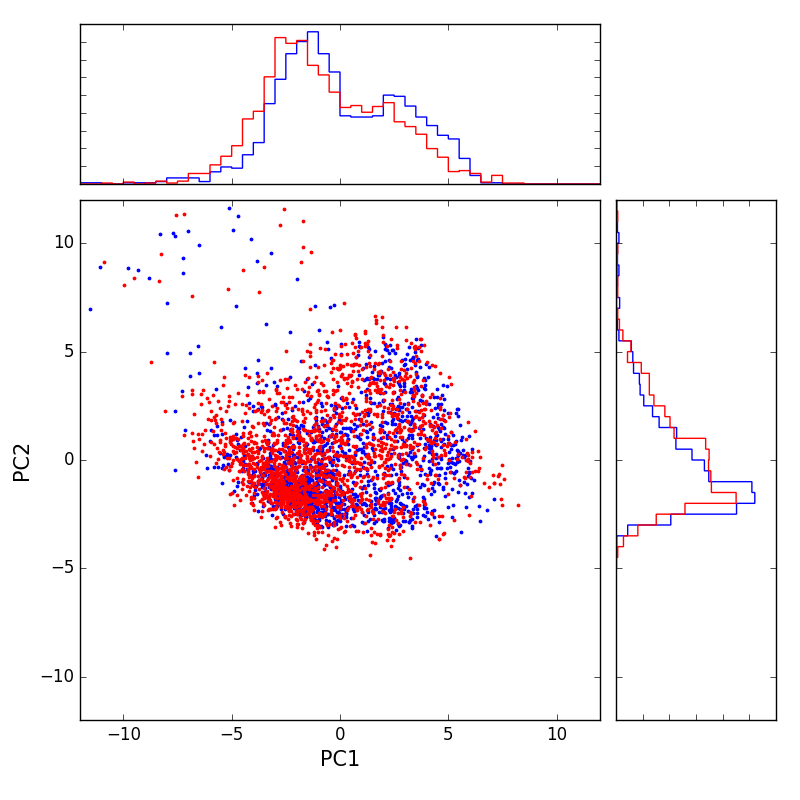}
  \includegraphics[width=2.0in]{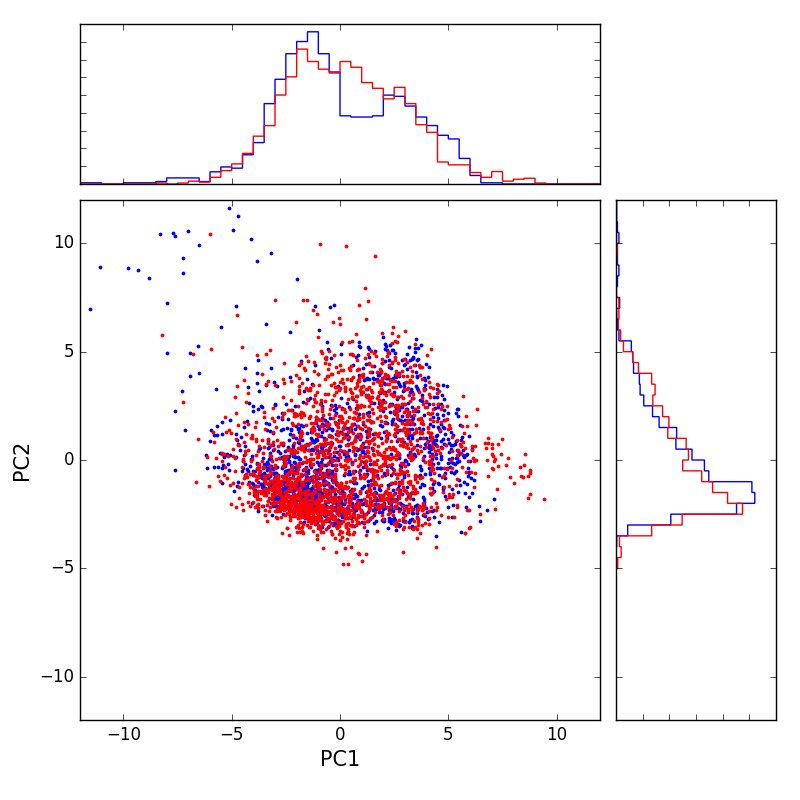}
  \includegraphics[width=2.0in]{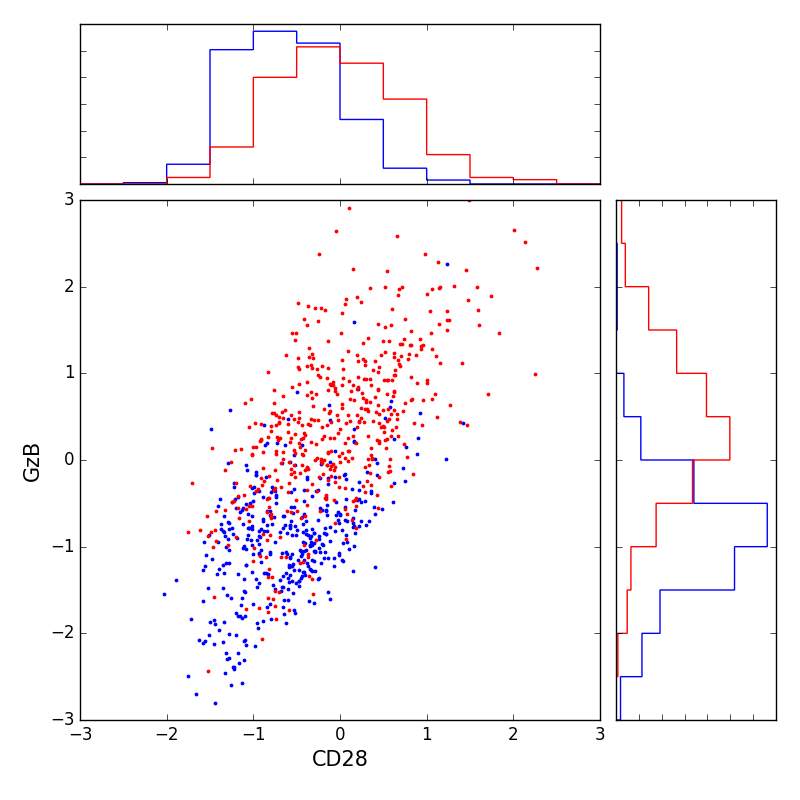}
  \includegraphics[width=2.0in]{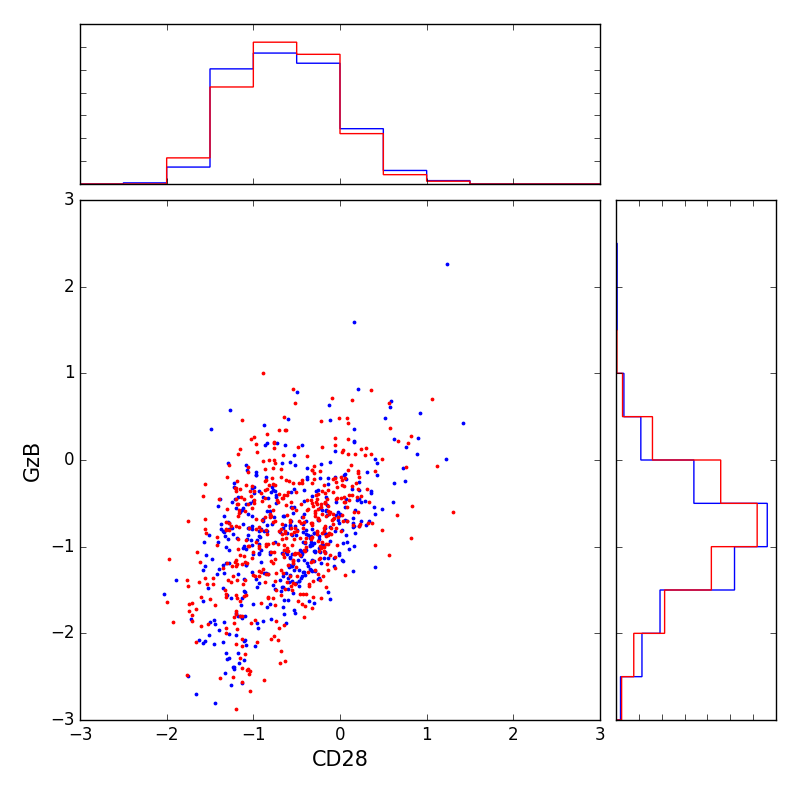}
  \includegraphics[width=2.0in]{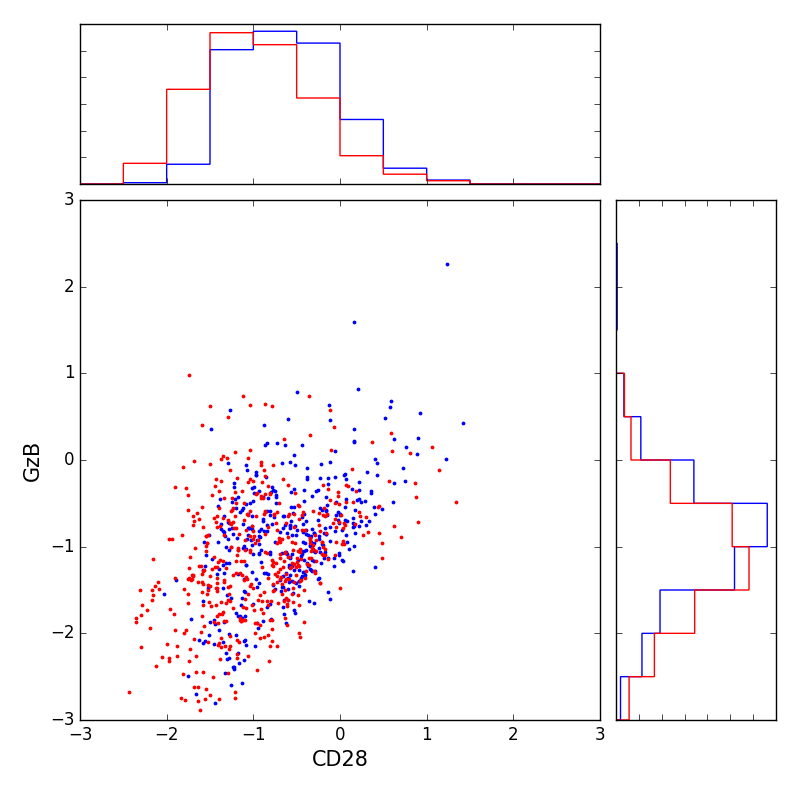}
  \caption{Indirect calibration of CyTOF data. Left: before calibration. Center: direct calibration. Right: indirect calibration. Top row: whole sample, projected onto the subspace of the first two principal components. Bottom row: CD8 sub-population in the (CD28,GzB) plane.}
 \label{fig:indirectCali}
 \end{figure}

We then mapped $p_1d_1$ to $p_1d_2$ through $N_{d_1}$, followed by $N_{p_2}$ and $N_{d_2}$ (while adjusting the means and variances at each point, to account for the fact that each of these nets was trained on a standardized source sample), and compared the resulting calibration to the direct calibration obtained by applying $N_{p_1}$ on $p_1d_1$. 
The results are presented in Figure~\ref{fig:indirectCali}.
As we can see, the indirect calibration is only slightly less accurate than direct calibration, and removes much of the batch effect. MMD between the source and target values support this observation: while before calibration the MMD is 0.69, it is 0.27 after direct calibration and 0.29 after indirect calibration. 

The success of removing much of the batch effect via indirect calibration in the above experiment implies that the biological state of the patient $p_1$ were not distorted by much during the propagation through the nets $N_{d_1}$ and $N_{d_2}$. This suggests that our MMD ResNets approach can be adapted for performing calibration in cases were replicates from a reference sample are measured in two batches and additional samples are measured only in one of the batches.

Suppose, for example, that in each day we run a CyTOF instrument to measure blood of a several (different) individuals, and in addition we also measure in each of these runs a replicate of a reference blood sample.
One can train a MMD-ResNet at each day $t$ to calibrate the reference blood sample to its distribution at day $0$. 
In addition, every replicate measured on day $t$ can be calibrated using (a different) MMD-ResNet to the reference sample at day $t$.  This way data from all days may be compared, by mapping all samples to coordinates of the reference sample at day 0.

%%%%%%%%%%%%%%%%%%%%%%%%%%%%%%%%%%%%%%%%%%%%%%%%%%%%%%%%%%%%%%%%%%%%%%%%%%%%%%%%%%%%%%%%%%%%%%%%%%%%%%%
%%%%%%%%%%%%%%%%%%%%%%%%%%%%%%%%%%%%%%%%%%%%%%%%%%%%%%%%%%%%%%%%%%%%%%%%%%%%%%%%%%%%%%%%%%%%%%%%%%%%%%%

\section{Related Work} \label{sec:relatedWork}

\citet{leek2010tackling} thoroughly discuss the importance of tackling batch effects and review several existing approaches for doing so. 

Bead normalization~\citep{finck2013normalization} is a specific normalization procedure for CyTOF. 
As we observed in Section~\ref{sec:experiments}, two CyTOF samples may significantly differ in distribution even after Bead normalization.
Warping~\citep{hahne2010per} is an approach for calibration of cytometry data where for each marker, the peaks of the marginal distribution in the source sample are (possibly non-linearly) shifted to match the peaks of the corresponding marginal in the target sample. 
We argue that warping can perhaps be performed by training MMD-ResNet for each single marker. 
The advantage of MMD-ResNet over a warping is that the former is multivariate, and can take into account dependencies, while the latter assumes that the joint distributions is a product of its marginals~\citep{finak2014high}. 

Surrogate variable Analysis~\citep{leek2007capturing} is a popular approach for batch effect adjustment, primarily in gene expression data. However, it is designed for supervised scenarios where labels representing the phenotype of each gene expression profile are provided, hence it is not directly applicable

MMD was used as a loss criterion for artificial neural networks in~\citep{li2015generative, dziugaite2015training}, where the goal was to learn a generative model that can transform standard input distributions (e.g. white noise) to a target distribution. To the best of our knowledge, MMD nets have not been applied to the problem of removal of batch effects, which is considered here.

%%%%%%%%%%%%%%%%%%%%%%%%%%%%%%%%%%%%%%%%%%%%%%%%%%%%%%%%%%%%%%%%%%%%%%%%%%%%%%%%%%%%%%%%%%%%%%%%%%%%%%%
%%%%%%%%%%%%%%%%%%%%%%%%%%%%%%%%%%%%%%%%%%%%%%%%%%%%%%%%%%%%%%%%%%%%%%%%%%%%%%%%%%%%%%%%%%%%%%%%%%%%%%%
\section{Discussion} \label{sec:discussion}
The problem of learning generative models has drawn much attention in the machine learning community recently.  Evaluation of such models, however, is not always fully clear.  Many recent works proposing generative models use Parzen window estimates for model evaluation. 
As~\cite{theis2015note} nicely point out, evaluation of generative models using Parzen windows is problematic; in our context, for example, suppose that the net maps the source points to the centers of mass of the target sample. Such a map will have high Parzen likelihood estimates, while clearly not calibrating the data well. MMD, which takes also into account the internal structure of the calibrated source  sample (term which is missing in Parzen estimates) might be more suitable for evaluation of the quality of the calibration.

In some of our experiments, which are not reported here, we found out that identifying cluster structure of the data might be a useful practice prior to applying MMD-ResNets in certain applications.
For instance, when one uses CyTOF to characterize Peripheral Blood Mononuclear Cells (PBMCs), the multi-marker cell distributions typically have separable clusters, corresponding to cell type sub-populations. 
While the relative proportion of different cell types in two replicate blood samples is expected to be invariant to the CyTOF machine, measuring these samples in two different runs in the same instrument or two different instruments often show noticeable differences between the cell type composition.
When the proportions of corresponding clusters differ between the source and target distributions, we do not expect that MMD-ResNet will account for that difference, as it computes a continuous map. 
In such cases, for example, it might be useful to use sub-sampling in order to match the relative proportions of each cell type between the source and the target samples.

%%%%%%%%%%%%%%%%%%%%%%%%%%%%%%%%%%%%%%%%%%%%%%%%%%%%%%%%%%%%%%%%%%%%%%%%%%%%%%%%%%%%%%%%%%%%%%%%%%%%%%%
%%%%%%%%%%%%%%%%%%%%%%%%%%%%%%%%%%%%%%%%%%%%%%%%%%%%%%%%%%%%%%%%%%%%%%%%%%%%%%%%%%%%%%%%%%%%%%%%%%%%%%%
\section{Conclusion} \label{sec:conclusion}
We presented a novel deep learning approach for non-linear removal of batch effects, based on residual networks, to match the distributions of the source and target samples.
We applied our approach to CyTOF and scRNA-seq and demonstrated impressive performance. To the best of our knowledge, such a performance on CyTOF data was never reported. Yet, our approach is general and can be applied to various data types.
To justify our approach, we showed that equivalent nets that lack the shortcut connections may distort the biological conditions manifested in the samples, while residual nets preserve them.
We also presented a novel approach for indirect calibration, which, to the best of our knowledge, is not performed elsewhere. 
It is based on an appealing property of using neural nets for calibration, which is the fact that the nets define a map, that can be later one applied to new data.

Lastly, despite the impressive experimental results presented here, a two sample test (say, a permutation test using MMD as a test statistic) will reject the hypothesis that the calibrated source sample has the same distribution as the target sample. 
Yet, in the same way that general deep learning techniques, operating on raw data outperform traditional algorithms tailored for specific data types and involving domain knowledge and massive pre-processing, we find our proposed approach and experimental results very promising and hope that they open new directions for removing batch effects in biological datasets.
For example, recent proposed experimental approaches to standardization~\citep{kleinsteuber2016standardization}, should provide an excellent source for application of MMD-ResNet for calibration.

%%%%%%%%%%%%%%%%%%%%%%%%%%%%%%%%%%%%%%%%%%%%%%%%%%%%%%%%%%%%%%%%%%%%%%%%%%%%%%%%%%%%%%%%%%%%%%%%%%%%%%%
%%%%%%%%%%%%%%%%%%%%%%%%%%%%%%%%%%%%%%%%%%%%%%%%%%%%%%%%%%%%%%%%%%%%%%%%%%%%%%%%%%%%%%%%%%%%%%%%%%%%%%%

\section*{Acknowledgement}
This research was partially funded by NIH grant 1R01HG008383-01A1 (Y.K.).
%%%%%%%%%%%%%%%%%%%%%%%%%%%%%%%%%%%%%%%%%%%%%%%%%%%%%%%%%%%%%%%%%%%%%%%%%%%%%%%%%%%%%%%%%%%%%%%%%%%%%%%
%%%%%%%%%%%%%%%%%%%%%%%%%%%%%%%%%%%%%%%%%%%%%%%%%%%%%%%%%%%%%%%%%%%%%%%%%%%%%%%%%%%%%%%%%%%%%%%%%%%%%%%

\bibliography{calibReferences_v5}
\bibliographystyle{apalike}

%%%%%%%%%%%%%%%%%%%%%%%%%%%%%%%%%%%%%%%%%%%%%%%%%%%%%%%%%%%%%%%%%%%%%%%%%%%%%%%%%%%%%%%%%%%%%%%%%%%%%%%
%%%%%%%%%%%%%%%%%%%%%%%%%%%%%%%%%%%%%%%%%%%%%%%%%%%%%%%%%%%%%%%%%%%%%%%%%%%%%%%%%%%%%%%%%%%%%%%%%%%%%%%

\clearpage

\appendix

%%%%%%%%%%%%%%%%%%%%%%%%%%%%%%%%%%%%%%%%%%%%%%%%%%%%%%%%%%%%%%%%%%%%%%%%%%%%%%%%%%%%%%%%%%%%%%%%%%%%%%%
%%%%%%%%%%%%%%%%%%%%%%%%%%%%%%%%%%%%%%%%%%%%%%%%%%%%%%%%%%%%%%%%%%%%%%%%%%%%%%%%%%%%%%%%%%%%%%%%%%%%%%%
\section{Specification of Markers in CyTOF Experiments}\label{app:markers}
Table\ref{tab:markers} provides the information about the 25 markers used in the CyTOF experiments in Section~\ref{sec:cytof}.
 \begin{table}[t]
\centering
\caption {Specification of the 25 markers used to characterize cell sub-populations in our CyTOF experiments.
Bead standards are embedded in each sample to allow Bead normalization.
Each Bead contains the four heavy metal isotopes labeled by 1 in the third column.}
\vskip 0.15in
\begin{tabular}{  c | c | c}
  \hline                        
  Isotope  & Marker  & Beads \\ \hline\hline      
  89Y    &  CD45    &       0\\ \hline
  142Nd  &  CD19    &       0\\ \hline
  143Nd  &  CD127   &       0\\ \hline
  145Nd  &  CD4     &       0\\ \hline
  146Nd  &  CD8a    &       0\\ \hline
  147Sm  &  CD20    &       0\\ \hline
  149Sm  &  CD25    &       0\\ \hline
  151Eu  &  CD278   &       1\\ \hline
  152Sm  &  TNFa    &       0\\ \hline
  153Eu  &  Tim3    &       1\\ \hline
  155Gd  &  CD27    &       0\\ \hline
  156Gd  &  CD14    &       0\\ \hline
  159Tb  & CCR7     &       0\\ \hline
  160Gd  & CD28     &       0\\ \hline
  161Dy  & CD152    &       0\\ \hline
  162Dy  & FOXP3    &       0\\ \hline
  164Dy  & CD45RO   &       0\\ \hline
  165Ho  & INFg     &       1\\ \hline
  166Er  & CD223    &       0\\ \hline
  167Er  & GzB      &       0\\ \hline
  170Er  &  CD3     &       0\\ \hline
  172Yb  & CD274    &       0\\ \hline
  174Yb  & HLADR    &       0\\ \hline
  175Lu  & PD1      &       1\\ \hline
  209Bi  & CD11b    &       0\\ \hline
  \end{tabular} 
  \label{tab:markers}
\end{table}

%%%%%%%%%%%%%%%%%%%%%%%%%%%%%%%%%%%%%%%%%%%%%%%%%%%%%%%%%%%%%%%%%%%%%%%%%%%%%%%%%%%%%%%%%%%%%%%%%%%%%%%
%%%%%%%%%%%%%%%%%%%%%%%%%%%%%%%%%%%%%%%%%%%%%%%%%%%%%%%%%%%%%%%%%%%%%%%%%%%%%%%%%%%%%%%%%%%%%%%%%%%%%%%
\section{Additional Plots for CyTOF calibration}\label{app:other3}
\begin{figure}
  \centering
  \includegraphics[width=2.5in]{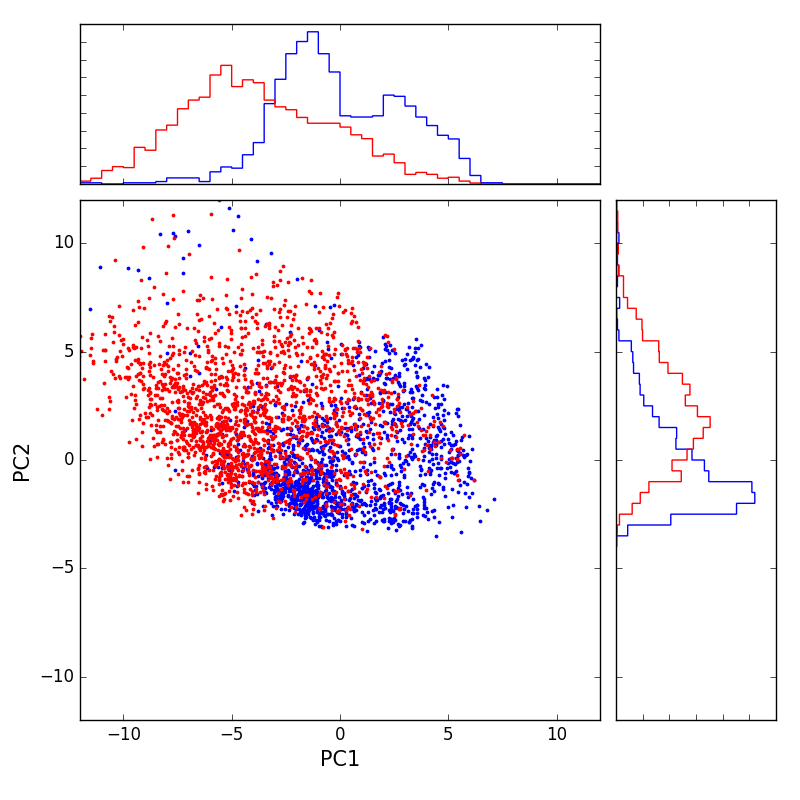}
  \includegraphics[width=2.5in]{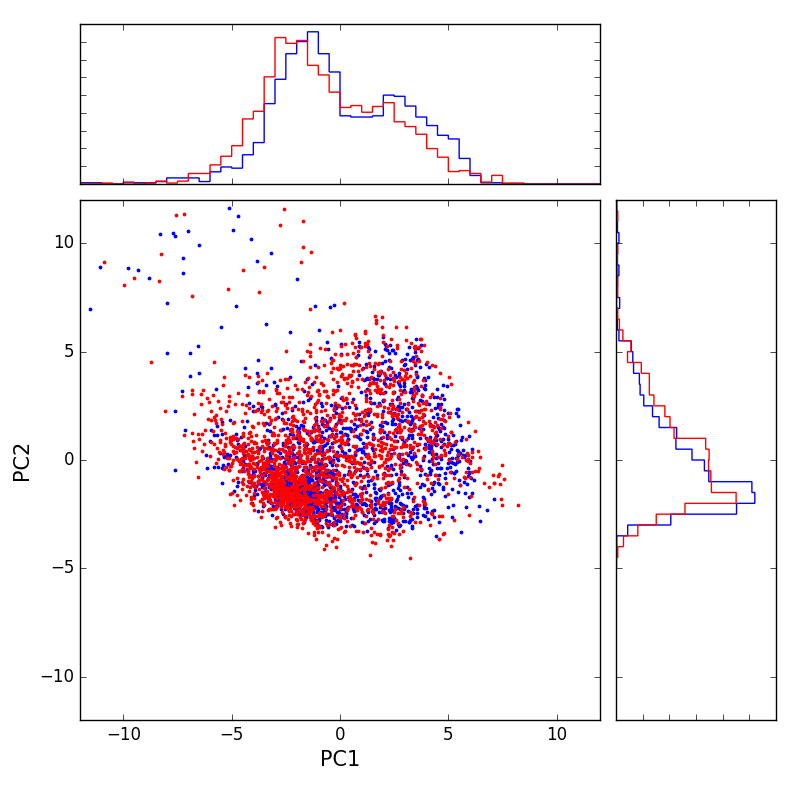}
  \includegraphics[width=2.5in]{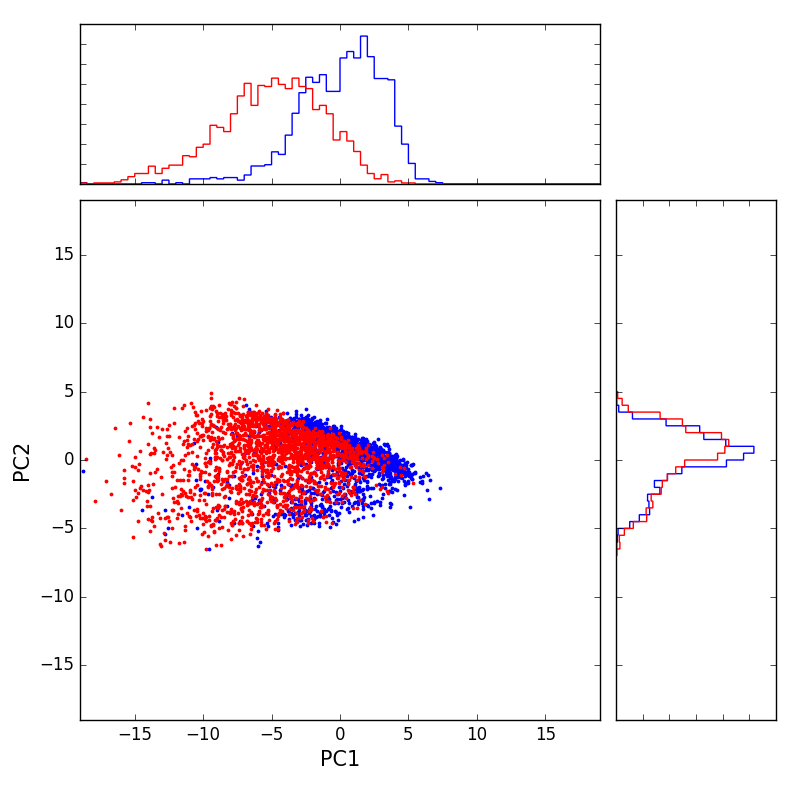}
  \includegraphics[width=2.5in]{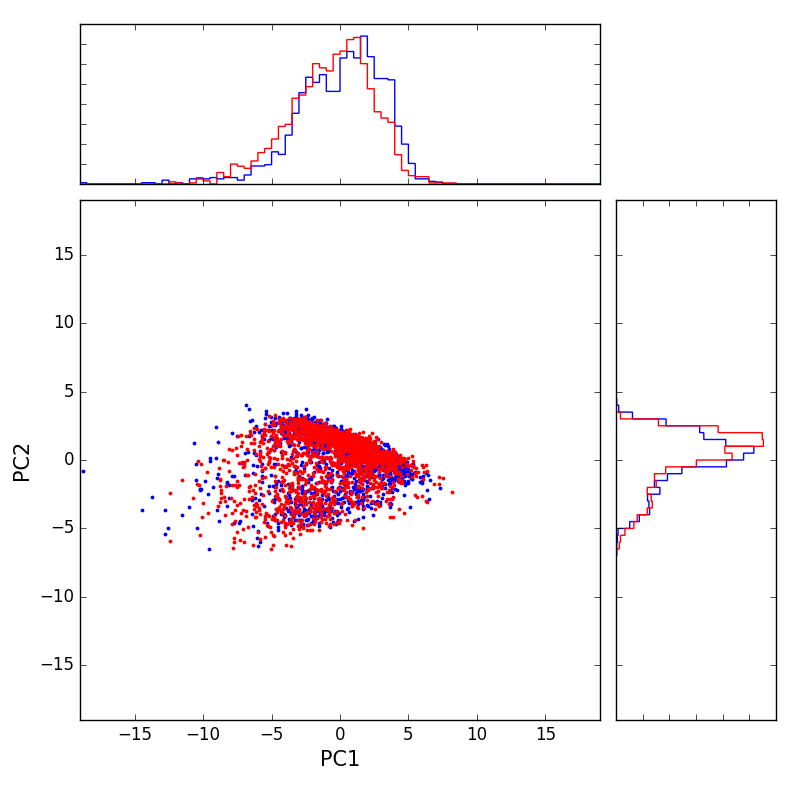}
  \includegraphics[width=2.5in]{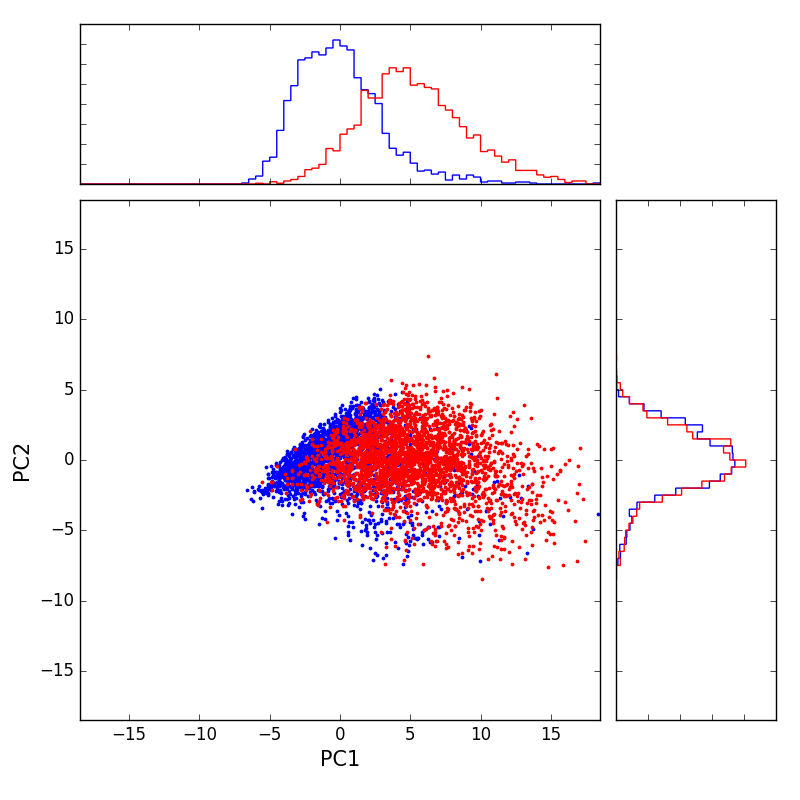}
  \includegraphics[width=2.5in]{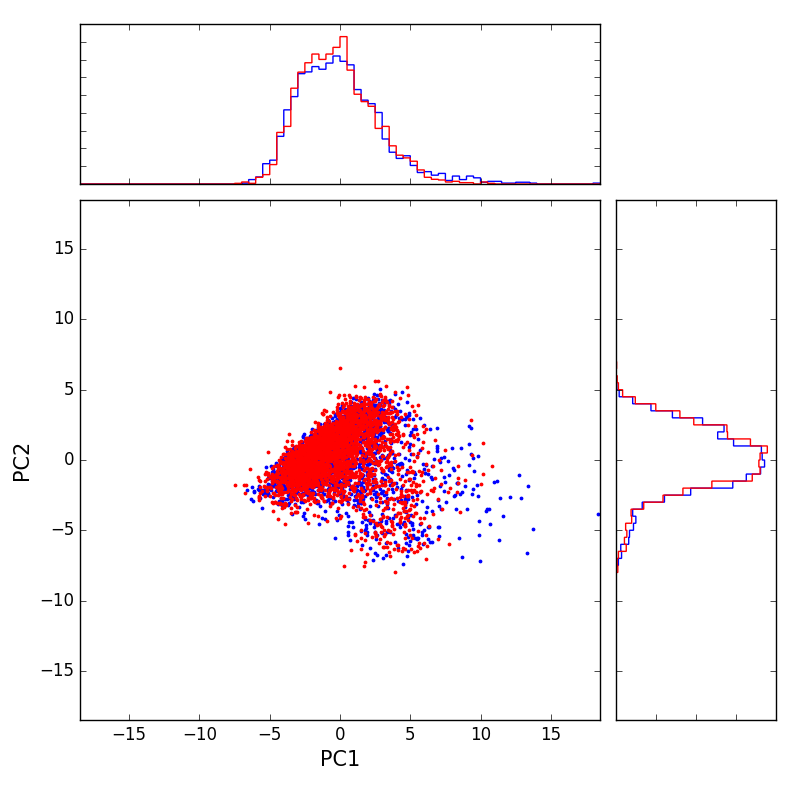}
  \caption{Calibration of CyTOF data, for each of the three source-target pairs not shown in Figure~\ref{fig:example}. Projection of the source (red) and target (blue) samples on the first two principal components of the target data. Left: before calibration. Right: after calibration.}
 \label{fig:other3}
 \end{figure}

Figure~\ref{fig:other3} shows the projection of the source and target samples onto the first two principal components of the target sample for the three additional source-target pairs not shown in Figure~\ref{fig:example}.

\begin{figure}
  \centering
  \includegraphics[width=2.0in]{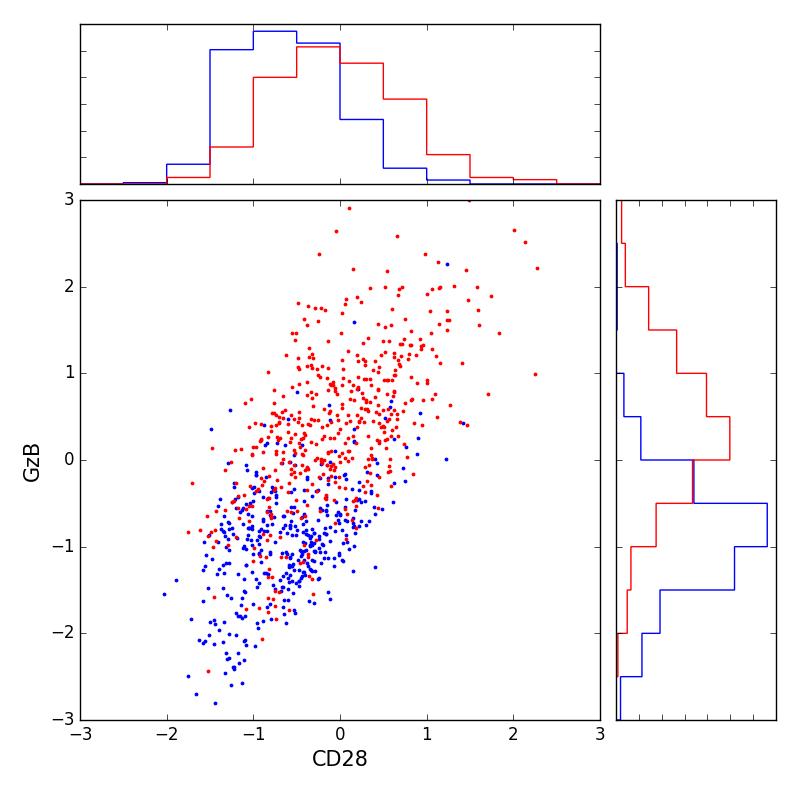}
  \includegraphics[width=2.0in]{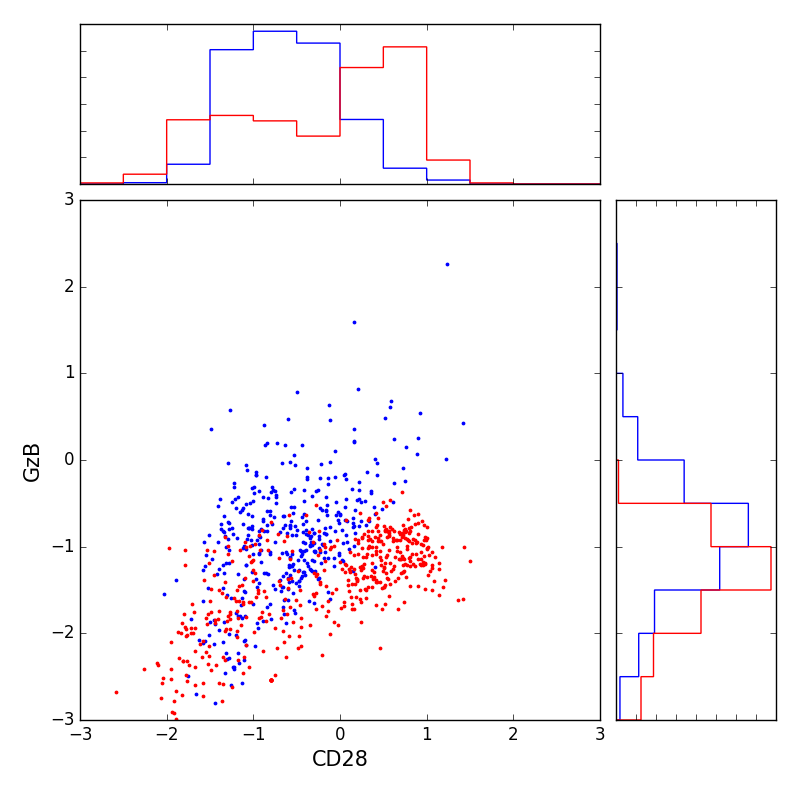}
  \includegraphics[width=2.0in]{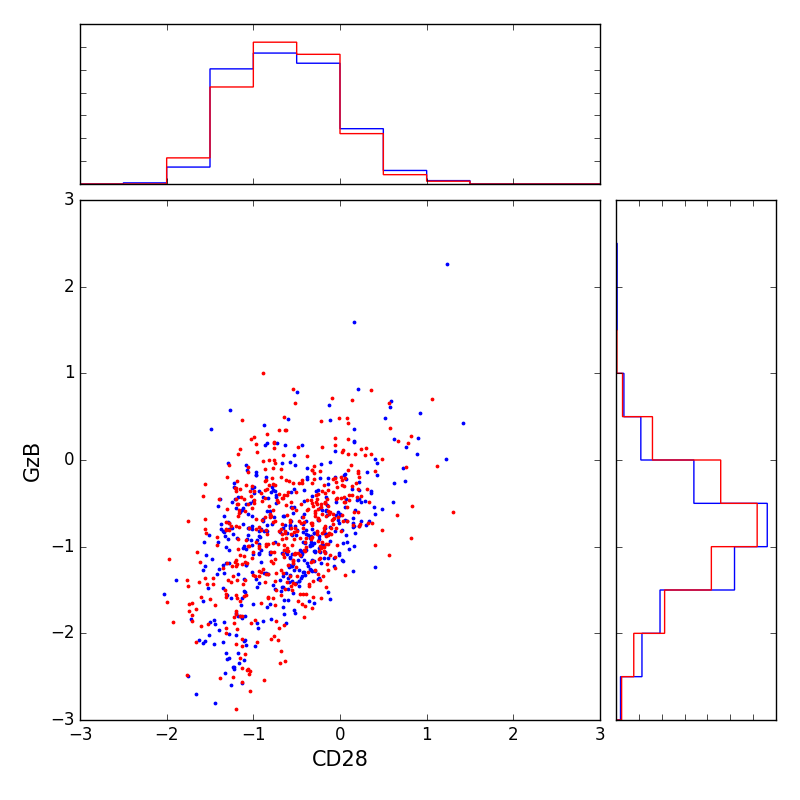}
  \includegraphics[width=2.0in]{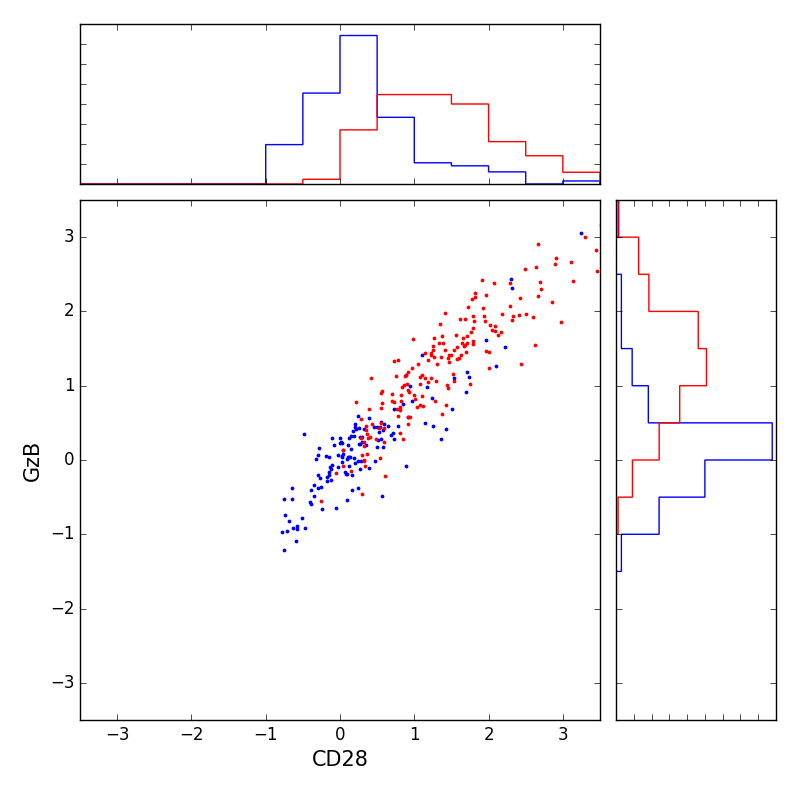}
  \includegraphics[width=2.0in]{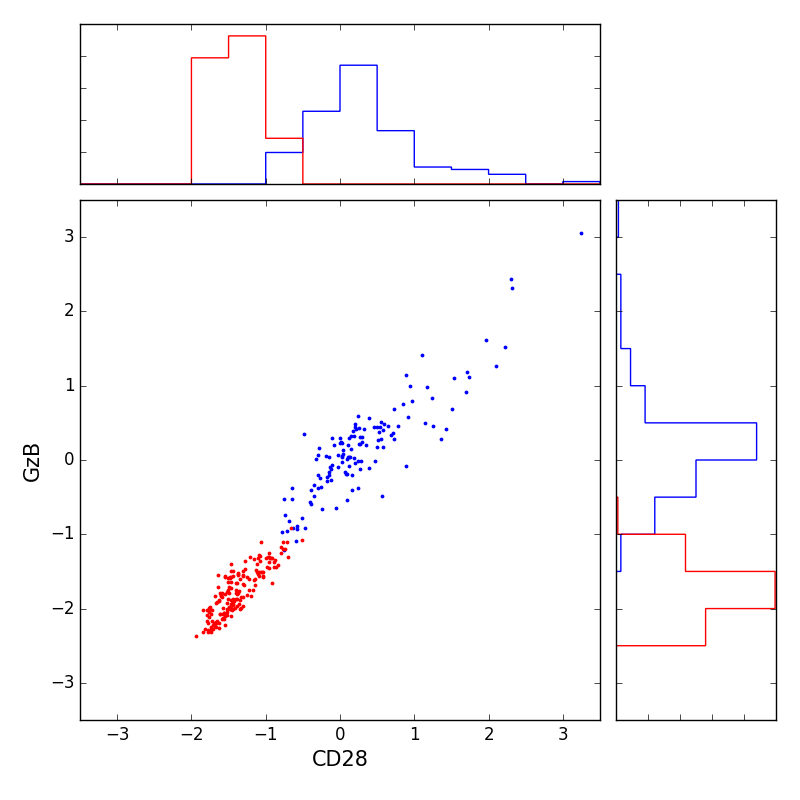}
  \includegraphics[width=2.0in]{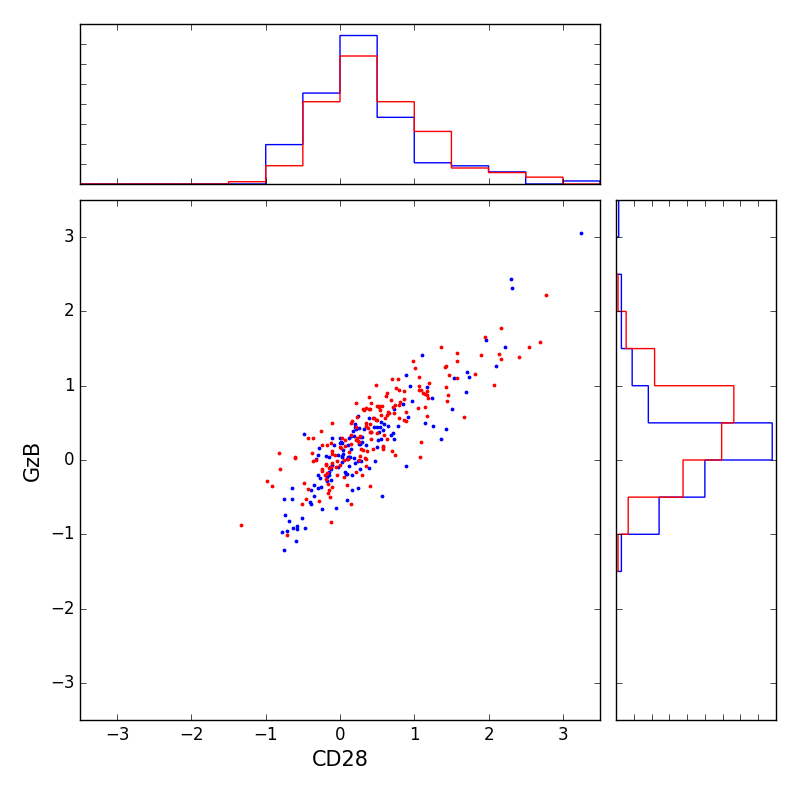}
  \includegraphics[width=2.0in]{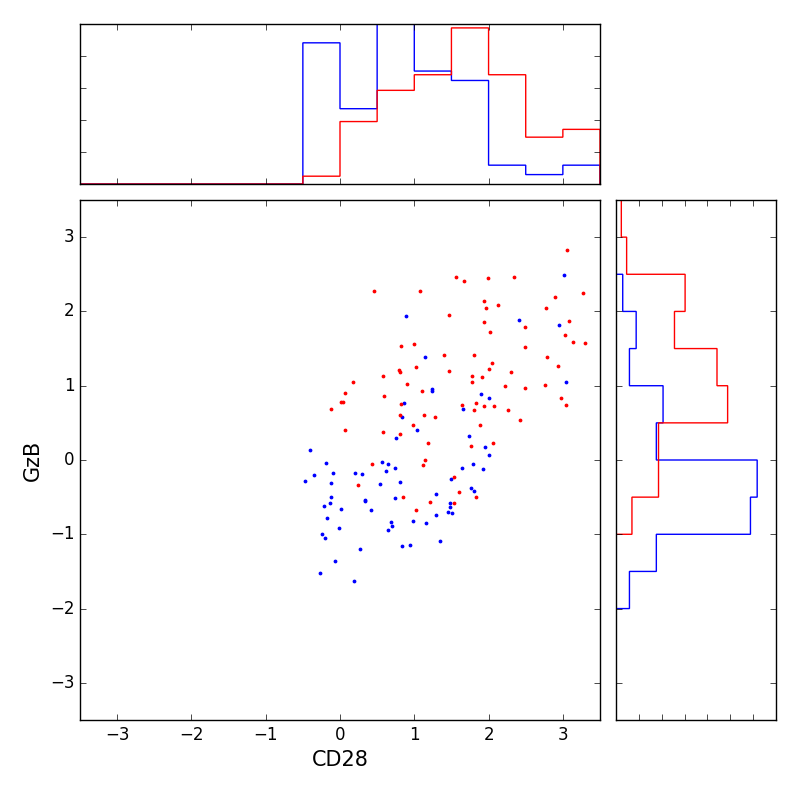}
  \includegraphics[width=2.0in]{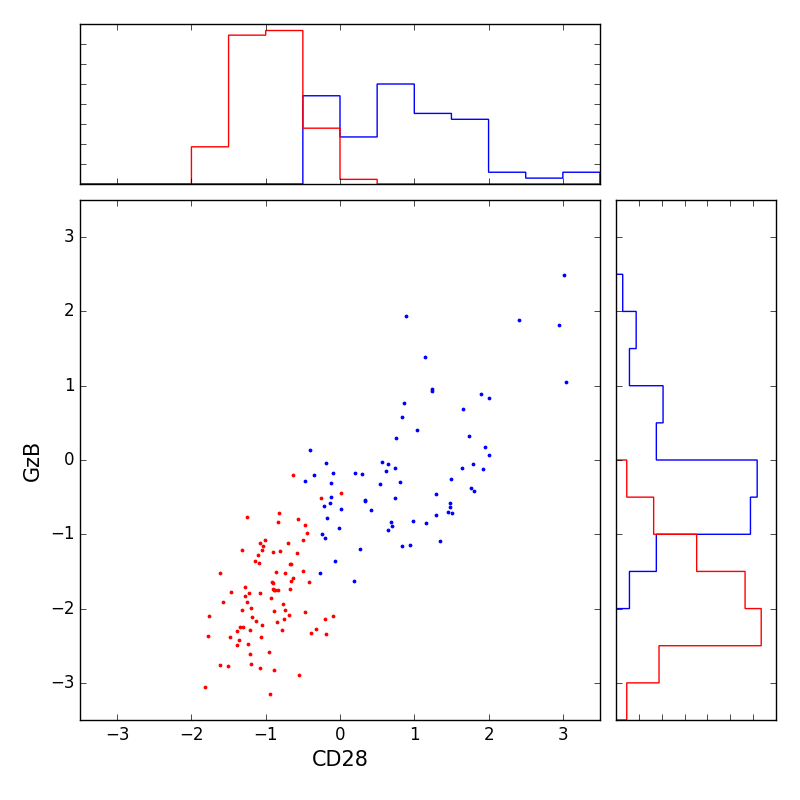}
  \includegraphics[width=2.0in]{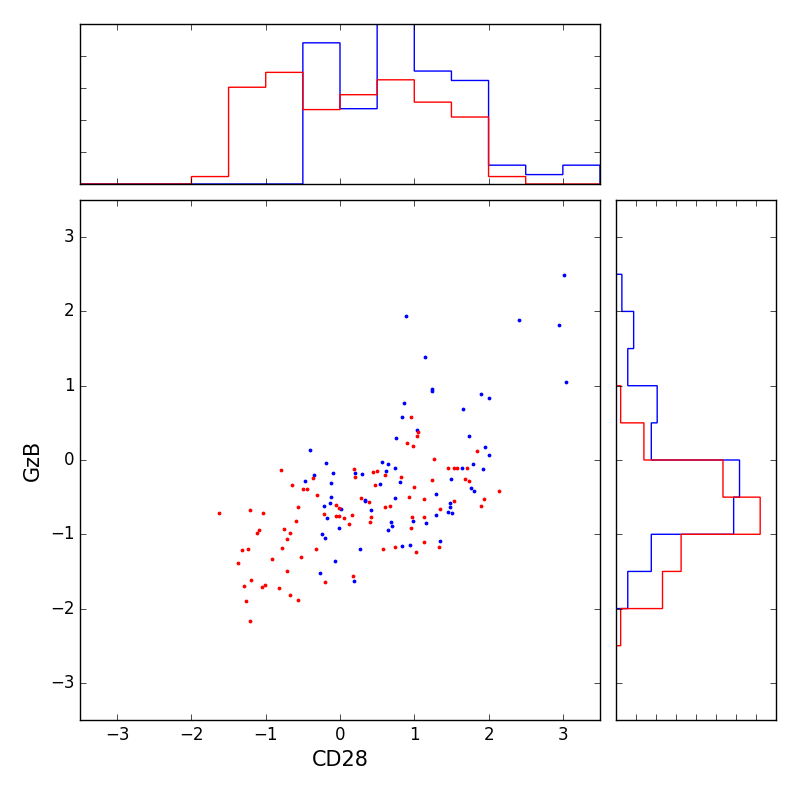}
  \caption{Calibration of CD8+T-cells sub-population in the (CD28,GzB) plane, for each of the three source-target pairs not shown in Figure~\ref{fig:cd8}. In each row the left plot corresponds to before calibration, the right to calibration using ResNet, and the center to calibration using an identical net, without shortcut connections and initialized in a standard fashion.}
 \label{fig:other3_cd8}
 \end{figure}

Figure~\ref{fig:other3_cd8} shows the projection of the CD8+ T-cell sub-population in the source and target data onto the first two principal components of the target sample for the three additional source-target pairs not shown in Figure~\ref{fig:cd8}.

%%%%%%%%%%%%%%%%%%%%%%%%%%%%%%%%%%%%%%%%%%%%%%%%%%%%%%%%%%%%%%%%%%%%%%%%%%%%%%%%%%%%%%%%%%%%%%%%%%%%%%%
%%%%%%%%%%%%%%%%%%%%%%%%%%%%%%%%%%%%%%%%%%%%%%%%%%%%%%%%%%%%%%%%%%%%%%%%%%%%%%%%%%%%%%%%%%%%%%%%%%%%%%%

\end{document}